\documentclass[11pt]{article}
\usepackage[preprint]{templet}
\usepackage{times}
\usepackage{multirow}
\usepackage{booktabs}
\usepackage{xspace}
\usepackage{siunitx}
\usepackage{amsmath,amssymb}
\usepackage{hyperref}
\usepackage{url}
\usepackage{pifont}
\newcommand{\xmark}{\ding{55}}
\newcommand{\cmark}{\ding{51}}
\usepackage{tabularx}
\usepackage{graphicx}
\usepackage{longtable,array}
\usepackage{xcolor}
\usepackage{colortbl}
\usepackage{arydshln}
\usepackage{listings}
\usepackage{adjustbox}
\usepackage[most]{tcolorbox}
\usepackage{enumitem}
\tcbuselibrary{listings}
\usepackage{longtable}
\definecolor{defblue}{rgb}{0.1843, 0.3333, 0.6}

\newcolumntype{K}[1]{>{\raggedright\arraybackslash}m{#1}}
\newcolumntype{Y}{>{\raggedright\arraybackslash}X}

\title{SketchJudge: A Diagnostic Benchmark for Grading Hand-drawn Diagrams with Multimodal Large Language Models}

\author{Yuhang Su, Mei Wang, Yaoyao Zhong, Guozhang Li, Shixing Li, Yihan Feng, Hua Huang \\
\textbf{School of Artificial Intelligence, Beijing Normal University, Beijing, China} \\
\textbf{yuhangsu@mail.bnu.edu.cn}}

\begin{document}

\maketitle

\begin{abstract}
While Multimodal Large Language Models (MLLMs) have achieved remarkable progress in \textit{visual understanding}, they often struggle when faced with the unstructured and ambiguous nature of human-generated sketches. This limitation is particularly pronounced in the underexplored task of \textit{visual grading}, where models should not only solve a problem but also diagnose errors in hand-drawn diagrams. Such diagnostic capabilities depend on complex \textit{structural}, \textit{semantic}, and \textit{metacognitive} reasoning. To bridge this gap, we introduce SketchJudge, a novel benchmark tailored for evaluating MLLMs as graders of hand-drawn STEM diagrams. SketchJudge encompasses 1,015 hand-drawn student responses across four domains: geometry, physics, charts, and flowcharts, featuring diverse stylistic variations and distinct error types. Evaluations on SketchJudge demonstrate that even advanced MLLMs lag significantly behind humans, validating the benchmark's effectiveness in exposing the fragility of current vision-language alignment in symbolic and noisy contexts. All data, code, and evaluation scripts are publicly available at \url{https://github.com/yuhangsu82/SketchJudge}.
\end{abstract}

\section{Introduction}

Multimodal Large Language Models (MLLMs) have made significant progress in tasks such as visual question answering, chart interpretation, and document understanding. However, as applications broaden, several fundamental and underexplored challenges have begun to surface.

The prevailing paradigm operates on a “perfect world” assumption, favoring clean and standardized images. However, hand-drawn sketches such as geometric constructions, free-body diagrams, and flowcharts shatter this idealized premise with their inherent ambiguity and stylistic diversity, yet remain indispensable to expressing complex reasoning in STEM education. Moreover, existing benchmarks almost exclusively cast models in the role of a \textit{solver}, while this design overlooks another critical role of a \textit{grader}. In settings like education, the core value is not to replace human thinking but to augment it, for instance, by acting as a grader that accurately identifies conceptual errors in a student's solution.

\begin{figure}[t]
  \centering
  \includegraphics[width=0.90\linewidth]{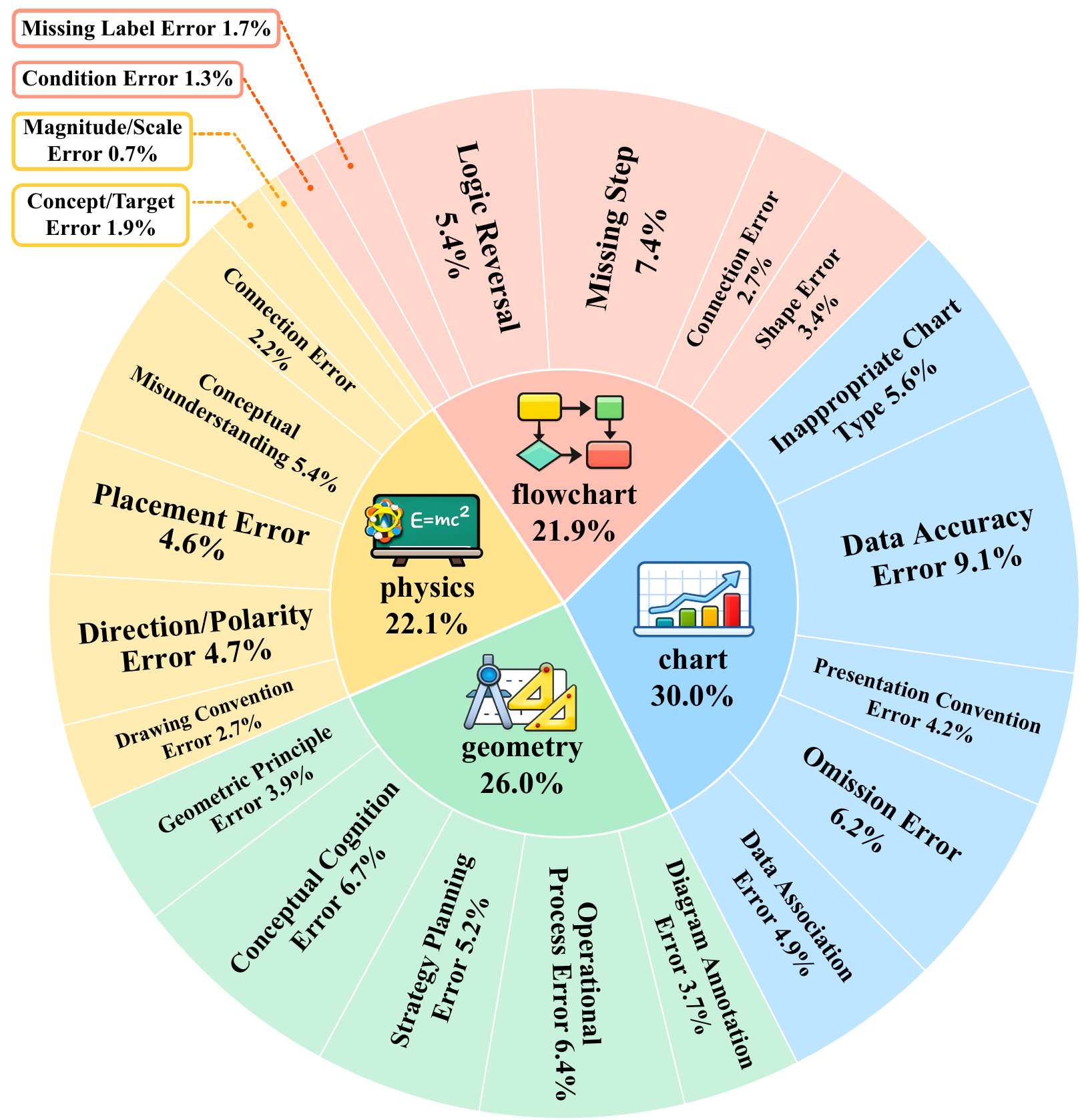}
  \caption{Distribution of annotated error types across four task domains in SketchJudge.}
  \label{fig:error_dist}
\end{figure}

Uniting this complex role with such unstructured input data defines a new frontier for MLLMs. The ability to evaluate diagrams is especially crucial in educational and assessment contexts. Acting as an effective grader for hand-drawn diagrams requires a multi-layered reasoning process that mirrors human cognition. At the perceptual level, this includes parsing sparse and ambiguous visual information despite variations in individual drawing styles. At the structural level, it involves understanding spatial relationships and logical connections. At the semantic level, it depends on mastery of domain-specific symbols and concepts. Finally, at the metacognitive level, it calls for assessing the overall plausibility and completeness of the solution. However, the lack of suitable benchmarks impedes research in this domain.


To address this gap, we introduce \textbf{SketchJudge}, a \textit{diagnostic} benchmark designed to evaluate the ability of MLLMs to grade hand-drawn diagrams and diagnose their limitations. The benchmark encompasses four domains—geometry, physics, charts, and flowcharts—comprising a total of 300 natural language questions and 1,015 hand-drawn student answers. The number of questions varies by domain, reflecting their different levels of complexity and annotation density. Each student's answer is paired with binary correctness labels and fine-grained error categories, with 5–7 domain-specific error types defined for each domain and validated by subject experts. Figure~\ref{fig:error_dist} provides an overview of the distribution of error types across task domains in SketchJudge. Compared to existing benchmarks like MathVista \citep{lu2023mathvista}, ChartQA \citep{masry2022chartqa}, and DocVQA \citep{mathew2021docvqa}, \textbf{SketchJudge} focuses on messy, symbolic, and structure-rich sketches that challenge shallow pattern-matching techniques.

Our main contributions are:
\begin{itemize}
    \item We introduce the first \textit{diagnostic} benchmark, to our knowledge, for \textbf{grading} hand-drawn diagrammatic answers in educational contexts, spanning four task types.
    \item We propose a fine-grained error taxonomy that enables structured diagnosis of model errors beyond simple correctness judgments.
    \item We present systematic experiments with several state-of-the-art MLLMs, revealing persistent limitations in diagram-level reasoning and offering insights for future research.
\end{itemize}

\section{Related Work}

\begin{table*}[t]
\centering
\small
\caption{
Comparison with closely related benchmarks. We mark whether a benchmark involves
\textit{freehand} diagram inputs, evaluates \textit{student-style} responses,
supports \textit{grading/judging} rather than solving, provides \textit{fine-grained}
error diagnosis, and includes an explicit \textit{reference} solution/diagram.
}
\label{tab:benchmark_comparison}
\setlength{\tabcolsep}{6pt}
\renewcommand{\arraystretch}{0.95}
    \begin{tabular}{lccccc}
    \toprule
    \textbf{Benchmark} &
    \textbf{Freehand} &
    \textbf{Student Resp.} &
    \textbf{Grading/Judge} &
    \textbf{Diag. Types} &
    \textbf{Ref. Provided} \\
    \midrule
    DocVQA~\citep{mathew2021docvqa} & \xmark & \xmark & \xmark & \xmark & \xmark \\
    ChartQA~\citep{masry2022chartqa} & \xmark & \xmark & \xmark & \xmark & \xmark \\
    MathVista~\citep{lu2023mathvista} & \xmark & \xmark & \xmark & \xmark & \xmark \\
    JudgeBench~\citep{tan2024judgebench} & \xmark & \xmark & \cmark & \xmark & \xmark \\
    HallusionBench~\citep{guan2024hallusionbench} & \xmark & \xmark & \cmark & \xmark & \xmark \\
    \midrule
    \textbf{SketchJudge (ours)} & \cmark & \cmark & \cmark & \cmark & \cmark \\
    \bottomrule
    \end{tabular}
\end{table*}


\paragraph{Multimodal Benchmarks for MLLMs.}
Evaluation of MLLMs has evolved from early QA-style benchmarks to broader frameworks that stress-test reasoning, planning, and text-rich understanding \citep{liu2024mmbench,yue2024mmmu,li2024seed}. More recent diagnostic benchmarks further probe cross-domain generalization and integrated capabilities \citep{ying2024mmt,hao2025can,yu2023mm}, while realism-oriented efforts incorporate high-resolution inputs or document-like scenarios \citep{li2024seed2,fu2024mme,wang2024mementos,zhang2024mme}. Despite this progress, most existing benchmarks still assume clean, standardized visual inputs and rarely evaluate grading-oriented behavior on noisy, freehand diagrams. As summarized in Table~\ref{tab:benchmark_comparison}, none of the closely related benchmarks jointly target \emph{freehand sketches}, \emph{diagram grading}, and \emph{fine-grained error diagnosis}. \textbf{SketchJudge} fills this gap by focusing on hand-drawn, structure-rich student diagrams and enabling diagnostic evaluation beyond answer matching.

\begin{figure*}[t]
    \centering
    \includegraphics[width=\linewidth]{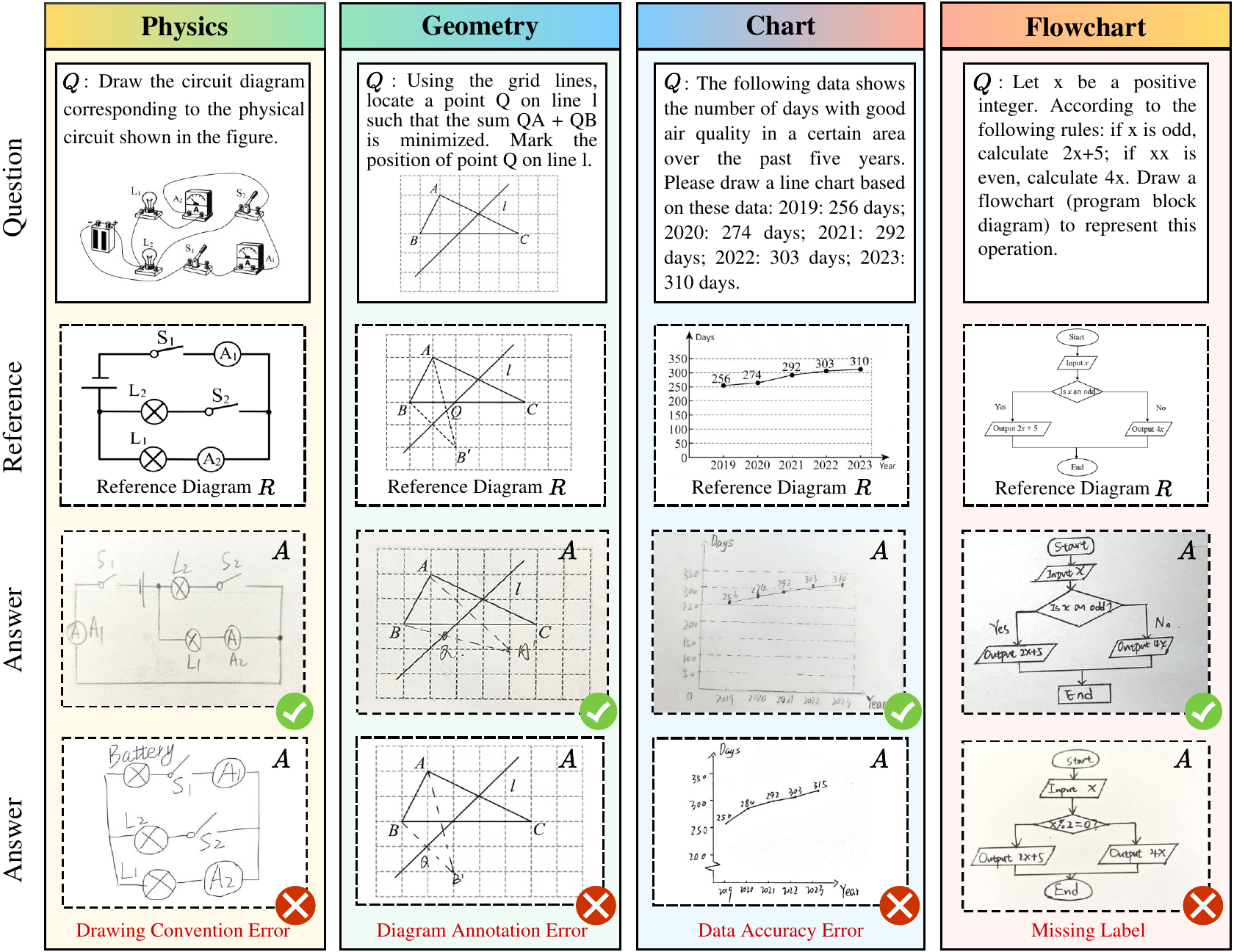}
    \caption{
        Representative SketchJudge instances across four domains, showing the question $Q$, reference diagram $R$, and student answers $A$ with correctness labels and error types.
    }
    \label{fig:sketchjudge_overview}
\end{figure*}

\paragraph{Evaluation Paradigms: Solver vs. Judge.}
Existing benchmarks have largely followed a “solver” paradigm, where MLLMs outputs are directly matched against ground-truth answers \citep{liu2024mmbench,yue2024mmmu,li2024seed}. In contrast, judge-based approaches position models as evaluators, assessing response quality along dimensions like correctness, helpfulness, and consistency \citep{zheng2023judging,tan2024judgebench,chen2024mllm}. This shift, initially explored in text-only evaluation, has recently been adopted in multimodal contexts such as hallucination detection and visual consistency assessment \citep{li2023evaluating,guan2024hallusionbench}. Such approaches are especially relevant for educational applications, where grading often requires diagnosing partial errors or structural flaws rather than assigning a simple right-or-wrong label \citep{xie2024grade}.

\paragraph{Data Realism and Educational Contexts.}
Many multimodal benchmarks use clean or synthetic visuals, such as standardized diagrams or programmatically rendered charts. Recent work has shifted toward more realistic inputs, including scanned documents and high-resolution natural images \citep{mathew2021docvqa,tanaka2024instructdoc,zhang2024mme}. However, these efforts remain largely separate from educational assessment, where student-generated responses are often messy, hand-drawn, and structurally irregular. In practice, educational datasets still predominantly focus on text-based answers---often multiple choice---in science and mathematics, offering limited evidence on how models handle freehand diagrammatic solutions \citep{lu2022learn,lu2023mathvista}. Very recent resources have begun exploring personalized error diagnosis and feedback generation \citep{zhang2025correctness,hsu2025mathedu}, but none target the grading of hand-drawn diagrams. \textbf{SketchJudge} fills this gap by introducing student-style sketches with fine-grained error annotations as a diagnostic probe for evaluating MLLMs in realistic educational contexts.

\section{Benchmark Design}

\subsection{Overview of SketchJudge}
\label{sec:sketchjudge_overview}

SketchJudge is a diagnostic benchmark for evaluating how MLLMs grade noisy, freehand diagrammatic answers, mirroring real classroom scoring scenarios. It spans four task domains—\textbf{geometry}, \textbf{physics}, \textbf{charts}, and \textbf{flowcharts}—with key dataset statistics summarized in Table~\ref{tab:dataset_stats}. Figure~\ref{fig:sketchjudge_overview} provides representative instances illustrating the benchmark setup. Each SketchJudge instance takes the form \{$Q$, $R$, $A$\}, consisting of a natural-language question ($Q$), a clean reference diagram ($R$), and one or more hand-drawn student answers ($A$). The freehand sketches introduce substantial ambiguity in perception and structure, posing a realistic challenge for grading-oriented MLLMs.

\setlength{\tabcolsep}{10pt}
\begin{table}[t]
  \centering
  \small
  \caption{Key statistics of the SketchJudge.}
  \label{tab:dataset_stats}

  \resizebox{\columnwidth}{!}{
  \begin{tabular}{p{0.56\linewidth} r}
    \toprule
    \textbf{Dataset Scale} & \textbf{Value} \\
    \midrule
    Total problems & 300 \\
    Total hand-drawn answers & 1015 \\
    Average answers per problem & 3.38 \\
    Total images (Q + R + A) & 1462(147/300/1015) \\
    \midrule
    \textbf{Input Modalities} & \\
    Questions containing images & 147 / 300 (49\%) \\
    Problems with reference diagrams & 300 (100\%) \\
    \midrule
    \textbf{Annotations} & \\
    Correct answers & 470 (46\%) \\
    Incorrect answers & 545 (54\%) \\
    Domain-specific error types & 23 \\
    \bottomrule
  \end{tabular}
  }
\end{table}
\setlength{\tabcolsep}{6pt}

\begin{figure*}[htbp]
    \centering
    \includegraphics[width=\linewidth]{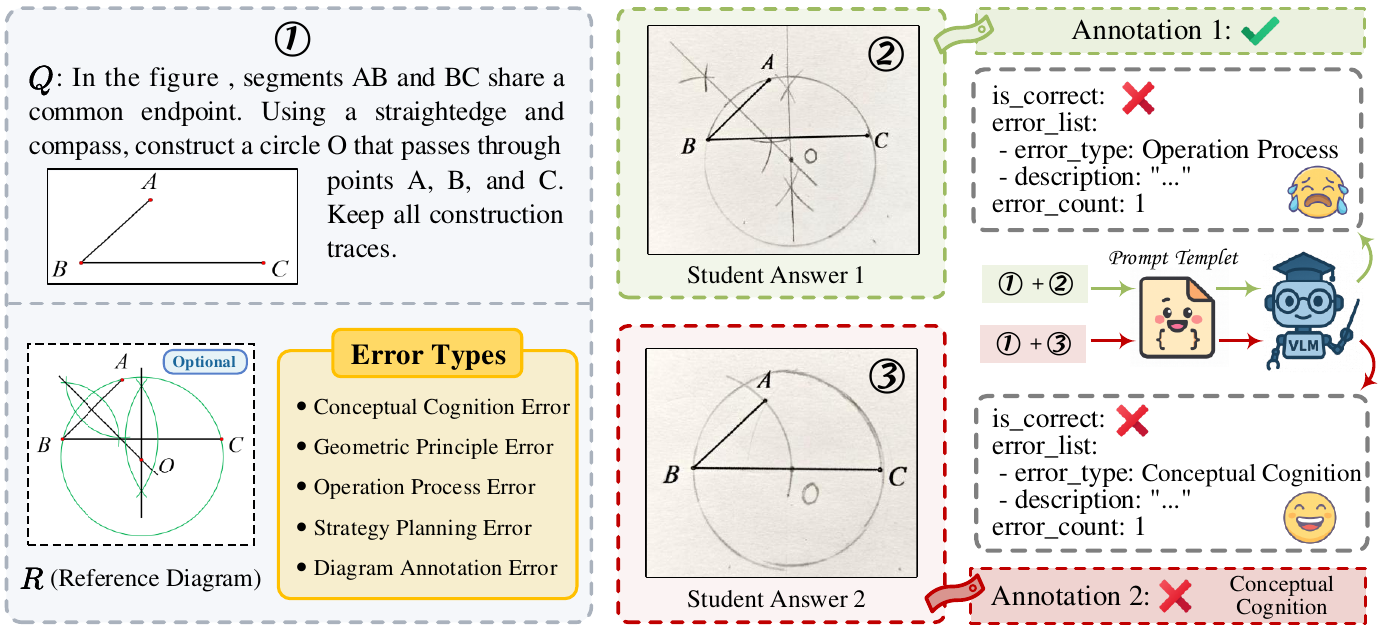}
    \caption{Illustration of the diagram grading task in SketchJudge.}
    \label{fig:framework}
\end{figure*}

To reflect common grading practices, SketchJudge defines a \emph{reference-based} setting (\textit{WithRef}) in which the model is provided with the reference diagram $R$. In addition, we introduce a complementary \emph{reference-free} setting (\textit{NoRef}) that withholds $R$, designed to stress-test a model’s ability to reason directly over the question $Q$ and student-drawn answer $A$ without reliance on a canonical solution. Figure~\ref{fig:framework} illustrates the grading formulation and expected outputs, including correctness judgments and fine-grained error-type diagnoses.

\subsection{Data Collection}
\label{sec:data_collection}

We collect SketchJudge with an emphasis on realism and diversity, aiming to reflect the variety of diagrammatic reasoning problems and classroom-style student submissions. Below we detail the data collection process for the three components of \{$Q$, $R$, $A$\} each instance.

\paragraph{Question.}
We curate four task categories: \textbf{geometry}, \textbf{physics}, \textbf{charts}, and \textbf{flowcharts}. Geometry and physics questions are primarily sourced from past exams and textbooks. For chart and flowchart tasks, we additionally include a small set of synthetic questions to control structural complexity and data diversity (e.g., varying chart types, scales, and flow structures), while keeping most questions real-world.

\paragraph{Reference.}
For each problem, we obtain a clean reference diagram ($R$) via one of three routes: (a) retrieving an official or widely accepted solution when available; (b) programmatically rendering chart references from the underlying data; or (c) manually authoring and validating a reference when no standard solution exists. All problem statements and references are verified by three independent raters, with disagreements resolved by consensus.

\paragraph{Answer.}
We collect \emph{1--9} student-drawn responses per question in either paper-based or digital form. We digitize the sketches as images, preserving realistic noise such as construction traces, uneven strokes, and minor skew. The sketches are contributed by 18 individuals, covering diverse drawing styles and common error patterns.

\subsection{Annotation Protocol}
\label{sec:annotation}

SketchJudge provides two levels of ground-truth annotations: a binary correctness label for each student sketch, and one or more fine-grained, domain-specific error types for each incorrect response, enabling diagnostic evaluation beyond correctness. To construct these taxonomies and apply them at scale, we adopt a two-stage annotation protocol.

\paragraph{Stage~1: Bottom-up taxonomy induction.}
Three annotators independently wrote natural-language \emph{error descriptions} for all incorrect sketches and cross-checked one another’s descriptions for clarity and consistency. From the resulting pool, we sampled 50 representative descriptions per domain to maximize coverage of distinct error phenomena, and prompted GPT-4o to cluster them into coherent categories, yielding a draft taxonomy. The generated categories were manually inspected to remove redundancy and resolve inconsistencies.

\paragraph{Stage~2: Expert refinement and taxonomy-based labeling.}
We refined the draft taxonomies through consultation with subject-matter experts, including frontline teachers in physics and math, adjusting category definitions and boundaries for educational plausibility. Using the finalized taxonomies, GPT-4o mapped each annotator-written error description to one or more error types, with the sketch withheld, producing initial labels that human annotators then verified and corrected when necessary; correct responses were labeled as error-free. Prompts used for taxonomy induction and labeling are provided in Appendix~\ref{app:prompts}.

\subsection{Challenges in Diagram Grading}
\label{sec:challenges}

Diagram grading in SketchJudge poses challenges that extend beyond conventional multimodal evaluation. Student-drawn diagrams exhibit substantial perceptual noise and stylistic variation, including uneven strokes, construction traces, and idiosyncratic drawing habits. Such variability complicates visual parsing and makes direct pixel- or feature-level matching unreliable.

Moreover, many tasks admit multiple valid diagrammatic solutions. These solutions may differ markedly in layout or orientation while remaining topologically and semantically correct. As a result, correctness cannot be determined by similarity to the reference diagram alone, but instead requires understanding the underlying task constraints and diagram semantics.

Beyond perceptual and structural variation, diagram grading demands fine-grained diagnostic judgment rather than coarse correctness assessment. As illustrated in Figure~\ref{fig:sketchjudge_overview}, sketches that deviate substantially from the reference may still be correct, whereas incorrect answers often contain subtle, localized errors, such as missing or incorrect labels, minor procedural mistakes, or small inaccuracies that alter semantic meaning. Distinguishing acceptable freehand imprecision from genuine conceptual or procedural errors requires models to reason about diagrammatic intent and apply domain knowledge, rather than relying on surface-level visual alignment.




\section{Experiments}
\label{sec:exp}

\subsection{Evaluation Settings}

We evaluate MLLMs on SketchJudge in a strictly zero-shot grading setup, following the \textit{WithRef} and \textit{NoRef} settings defined in Section~\ref{sec:sketchjudge_overview}. In our main experiments, we report \textit{WithRef} results, and provide \textit{NoRef} results and detailed comparisons in Section~\ref{sec:ablation_ref} and Appendix~\ref{app:full_results}. All models are evaluated with deterministic decoding in all settings (temperature~=~0).

\subsubsection{Evaluation Metrics}

\paragraph{Binary grading.}
Correctness prediction is formulated as a binary classification task. We report \textbf{Acc}, alongside \textbf{FNR} and \textbf{FPR} to diagnose over-strict and over-lenient tendencies. This allows us to disentangle raw performance from systematic grading bias.

\paragraph{Error-type recognition.}
For answers judged incorrect, we further evaluate recognition of fine-grained error types. Domain-specific taxonomies are unified into a namespaced global label set. Performance is reported using the example-based F1 (\textbf{eb$F_1$})~\cite{nam2017maximizing}, which measures the overlap between the predicted and gold error sets for each instance and averages over instances. We compute \textbf{eb$F_1$} only on instances where both the gold annotation and the model prediction indicate incorrectness. Detailed metric definitions are provided in Appendix~\ref{app:metrics}.

\subsubsection{Models}
We evaluate 16 state-of-the-art MLLMs, including both open-source and proprietary systems. All models are evaluated using identical prompts across domains, without any task-specific tuning. The full prompt templates are provided in Appendix~\ref{app:prompts}. A complete list of models and their detail information is provided in Appendix~\ref{app:models}.

\begin{table*}[t]
  \centering
  \caption{Overall leaderboard on SketchJudge. \textbf{Bold} and \underline{underline} indicate the best and second-best scores among models. Human performance is measured on a stratified subset of 160 student answers (40 per domain, balanced over correct/incorrect), graded by three non-expert raters following the same instructions; it is intended as a scalable baseline rather than an expert upper bound. All model results are computed on the full test set.}

  \label{tab:main_results}
  \adjustbox{max width=\textwidth}{
  \begin{tabular}{lcccccccccc}
    \toprule
    \multirow{2}{*}{\textbf{Model}} & \multicolumn{2}{c}{\textbf{Physics}} & \multicolumn{2}{c}{\textbf{Geometry}} & \multicolumn{2}{c}{\textbf{Chart}} & \multicolumn{2}{c}{\textbf{Flowchart}} & \multicolumn{2}{c}{\textbf{overall}}\\
    \cmidrule(lr){2-3}\cmidrule(lr){4-5}\cmidrule(lr){6-7}\cmidrule(lr){8-9}\cmidrule(lr){10-11}
    & Acc $\uparrow$ & eb$F_1 \uparrow$
    & Acc $\uparrow$ & eb$F_1 \uparrow$
    & Acc $\uparrow$ & eb$F_1 \uparrow$
    & Acc $\uparrow$ & eb$F_1 \uparrow$
    & Acc $\uparrow$ & eb$F_1 \uparrow$\\
    \midrule
    Random choice & 50.00 & -- & 50.00 & -- & 50.00 & -- & 50.00 & -- & 50.00 & -- \\
    \rowcolor{defblue!15}Human & 81.67 & 63.83 & 83.33 & 49.25 & 85.00 & 76.13 & 83.33 & 76.24 & 83.33 & 66.42 \\
    \hline
    \hline
    \multicolumn{11}{c}{\textit{\cellcolor{gray!10}Open-source models}} \\
    Llama-4-Scout-17B-16E-Instruct & 63.52 & 18.29 & 68.58 & 32.01 & 70.73 & 59.76 & 73.50 & 42.66 & 69.16 & 39.66 \\
    Gemma-3-4B-it & 55.36 & 25.05 & 51.34 & 21.27 & 55.40 & 43.94 & 51.28 & 38.67 & 53.40 & 31.66 \\
    Gemma-3-27B-it & 58.37 & 36.96 & 65.13 & 34.81 & 63.07 & \underline{74.08} & 68.38 & 54.64 & 63.74 & 49.28\\
    Qwen2.5-VL-3B-Instruct & 53.65 & 22.05 & 54.02 & 26.22 & 56.45 & 44.66 & 50.43 & 30.42 & 53.79 & 31.62 \\
    Qwen2.5-VL-7B-Instruct & 60.94 & 28.53 & 54.41 & 28.81 & 64.46 & 48.50 & 68.80 & 33.33 & 62.07 & 34.00 \\
    Qwen2.5-VL-72B-Instruct & 57.94 & 29.34 & 59.00 & 23.36 & 74.91 & 66.80 & 72.65 & 44.01 & 66.40 & 39.28 \\
    Qianfan-VL-8B & 45.92 & 48.15 & 52.49 & 22.22 & 70.03 & 32.49 & 65.81 & 38.82 & 59.01 & 32.40 \\
    Qianfan-VL-70B & 57.76 & 39.12 & 60.54 & 20.16 & 67.60 & 61.00 & 73.50 & 38.48 & 64.89 & 39.14 \\
    GLM-4.6V & 60.94 & 46.45 & 62.07 & 37.14 & 76.22 & 68.50 & 76.92 & 61.32 & 69.16 & 53.35 \\
    \hdashline
    \multicolumn{11}{c}{\textit{\cellcolor{gray!10}Closed-source models}} \\
    ERNIE-4.5-Turbo-VL & 60.09 & 39.03 & 66.28 & 25.94 & 72.13 & 51.89 & 66.24 & 37.34 & 66.50 & 38.56 \\
    Mistral-Large-3 & 57.08 & 29.69 & 64.37 & 32.16 & 68.99 & 56.06 & 75.21 & 49.07 & 66.50 & 41.96 \\
    Claude-3.7-Sonnet & 67.81 & \underline{48.33} & 68.97 & 36.44 & 75.61 & 70.59 & \underline{79.49} & 63.85 & 73.00 & 57.11 \\
    Doubao-Seed-1.6-Vision & 66.09 & 46.97 & 63.22 & 35.80 & 77.35 & 72.09 & 66.67 & 64.08 & 68.67 & 54.51 \\
    Gemini-2.5-Flash & 71.67 & \textbf{51.69} & \underline{77.04} & \textbf{45.98} & \underline{80.14} & 66.16 & \textbf{81.62} & \underline{66.90} & \underline{77.74} & \underline{58.30} \\
    o3 & \textbf{73.82} & 44.14 & 72.41 & 43.57 & 78.05 & 66.11 & 73.50 & 65.84 & 74.58 & 55.91 \\
    GPT-5 & \underline{73.39} & 46.83 & \textbf{77.78} & \underline{45.85} & \textbf{82.58} & \textbf{76.64} & 79.06 & \textbf{70.05} & \textbf{78.42} & \textbf{61.13} \\
    \bottomrule
  \end{tabular}
  }
\end{table*}

\subsection{Main Results}
\label{sec:mainresults}

Table~\ref{tab:main_results} reports the main leaderboard results on SketchJudge, showing that diagram grading remains challenging for current MLLMs. Human raters achieve an overall accuracy of 83.33\%, whereas the best-performing model (GPT-5) reaches 78.42\%, followed by Gemini-2.5-Flash (77.74\%) and o3 (74.58\%). Most open-source models obtain substantially lower overall accuracy, typically in the 53\%--69\% range. Since the human baseline is produced by independent non-expert raters following the same instructions as models, it reflects a \emph{scalable} grading setting rather than expert-level performance. We therefore treat it as a practical reference point for model comparison, complementing expert-validated ground-truth annotations.

We further observe a clear stratification between closed-source and open-source models in binary grading accuracy. Closed-source models generally deliver stronger and more consistent performance across domains, with multiple models exceeding 73\% overall accuracy (e.g., Claude-3.7-Sonnet, Gemini-2.5-Flash, o3, GPT-5). In contrast, open-source models exhibit both lower accuracy and larger cross-domain variance, even at larger scales. This gap suggests that robust grading requires reliable vision--symbol alignment and cross-domain generalization, which remain hard to match with current open-source MLLMs.

Beyond correctness, SketchJudge evaluates fine-grained diagnostic capability via error-type recognition. Human raters achieve an overall eb$F_1$ of 66.42, while the best models approach but do not match this level, with GPT-5 obtaining 61.13 and Gemini-2.5-Flash reaching 58.30. Interestingly, the strongest model differs across domains, indicating that grading performance is domain-sensitive and not yet uniformly solved. Overall, these findings support SketchJudge’s diagnostic focus and motivate future work on structured reasoning and supervision strategies that better ground error diagnoses in diagram semantics.

\subsection{Diagnostic Analyses}
\label{sec:diagnostic}

Beyond the aggregate results in Table~\ref{tab:main_results}, we also conduct diagnostic analyses to better understand how models behave as diagram graders, focusing on strictness/leniency (FNR/FPR), reliance on reference diagrams (\textit{WithRef} vs.\ \textit{NoRef}), and fine-grained error-type recognition. In addition, we report a broader range of diagnostic analyses and controlled ablations in Appendix~\ref{app:full_results}.

\subsubsection{Strictness and Leniency}
\label{sec:strict_lenient}

\begin{figure}[htbp]
    \centering
    \includegraphics[width=\linewidth]{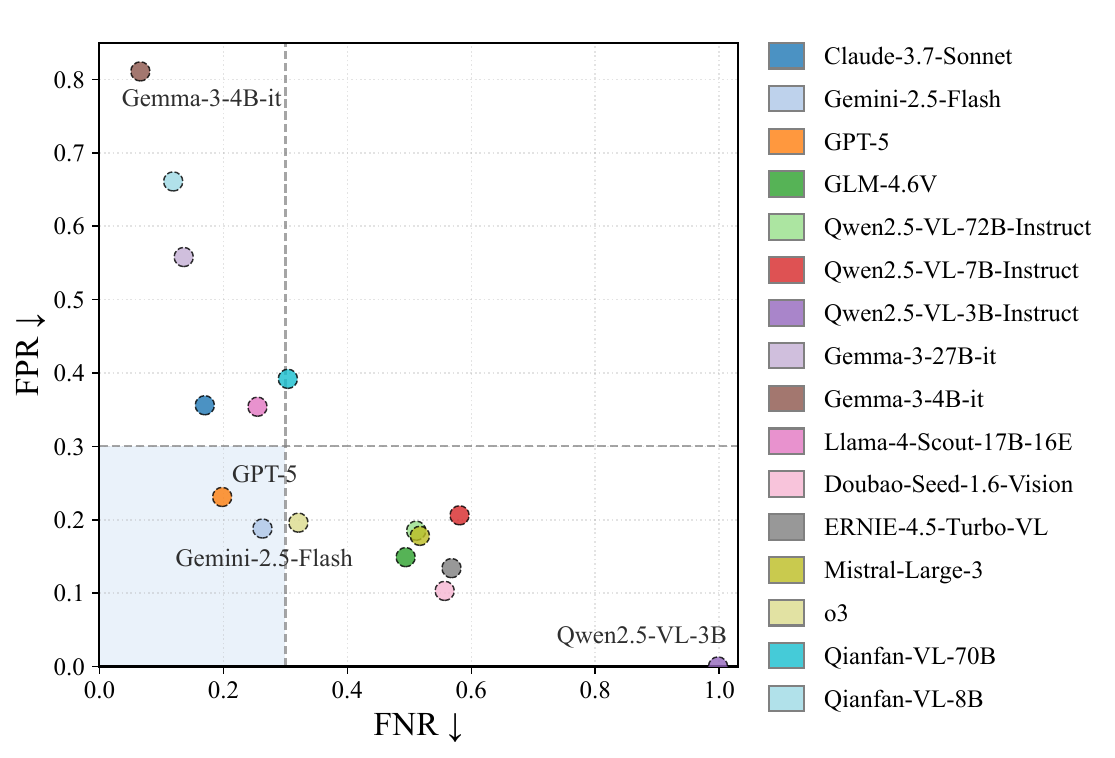}
    \caption{Strictness and leniency across models.}
    \label{fig:fnr_fpr}
\end{figure}

We analyze grading bias using the false-negative rate (FNR) and false-positive rate (FPR). Figure~\ref{fig:fnr_fpr} positions all evaluated models in the FNR--FPR space, where the lower-left region indicates more balanced grading.

Figure~\ref{fig:fnr_fpr} shows substantial variation in grading tendencies. Several smaller open-source models are strongly \emph{over-strict}, with very high FNR that rejects many correct sketches (e.g., Qwen2.5-VL-3B-Instruct). Others are comparatively \emph{over-lenient}, exhibiting low FNR but much higher FPR, suggesting they may accept plausible-looking yet incorrect diagrams (e.g., Gemma-3-4B-it). Stronger closed-source models such as GPT-5 and Gemini-2.5-Flash lie closer to the balanced region, but still make nontrivial errors in both dimensions. Overall, reliable diagram grading requires tolerating freehand variation while detecting conceptual errors.

\subsubsection{Effect of Reference Answers}
\label{sec:ablation_ref}

\begin{figure}[htbp]
    \centering
    \includegraphics[width=\linewidth]{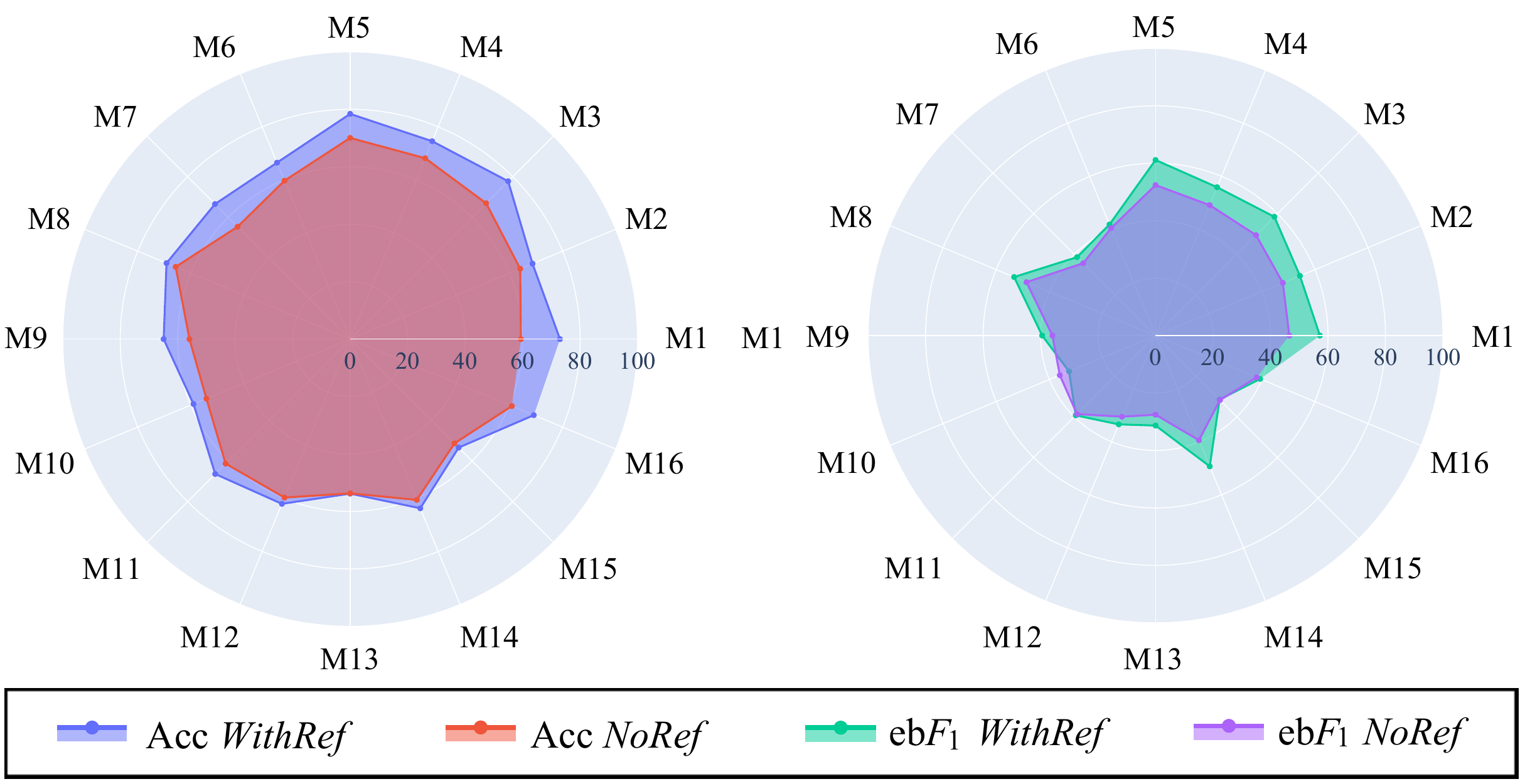}
    \caption{ Effect of reference diagrams across models. (Left) Accuracy comparison between \textit{WithRef} and \textit{NoRef} settings. (Right) Example-based F1 comparison under the same conditions.}
    \label{fig:ref_effect_radar}
\end{figure}

We quantify the effect of reference diagrams by comparing model performance under \textit{WithRef} and \textit{NoRef}. Figure~\ref{fig:ref_effect_radar} visualizes per-model changes in binary grading accuracy and error-type recognition (eb$F_1$); model indices follow Appendix~\ref{app:full_results}.

Providing a reference diagram improves accuracy for most models, suggesting that an explicit canonical solution helps align judgments with task constraints. In contrast, the effect on eb$F_1$ is less uniform: some models show clear improvements in error-type recognition, while others remain nearly unchanged. This gap suggests that reference availability primarily aids \emph{correctness verification}, whereas assigning fine-grained error types still requires deeper diagram-level interpretation beyond reference-based matching.

\subsubsection{Effect of Prompt Design}
\label{sec:prompt_design}

To test whether prompting can improve diagram grading, we compare three prompt styles: (i) a \textbf{Baseline} prompt with minimal instructions, (ii) a \textbf{Rubric} prompt that enumerates grading criteria and error types, and (iii) a \textbf{Chain-of-Thought (CoT)} prompt that encourages step-by-step comparison before making a final judgment.

\begin{table}[htbp]
  \centering
  \caption{Effect of prompt variants on grading performance.}
  \label{tab:prompt_main}

  \resizebox{\columnwidth}{!}{
  \begin{tabular}{llcccc}
    \toprule
    \textbf{Model} & \textbf{Prompt} & Acc & eb$F_1$ & FNR & FPR \\
    \midrule
    \multirow{3}{*}{Gemini-2.5-Flash}
        & Baseline & \textbf{77.74} & \textbf{58.30} & \textbf{0.263} & 0.188 \\
        & Rubric & 76.95 & 54.17 & 0.289 & 0.180 \\
        & CoT & 76.16 & 55.68 & 0.351 & \textbf{0.141} \\
    \midrule
    \multirow{3}{*}{GLM-4.6V}
        & Baseline & 69.16 & 53.35 & 0.494 & \textbf{0.149} \\
        & Rubric & \textbf{70.15} & \textbf{54.13} & \textbf{0.385} & 0.224 \\
        & CoT & 64.63 & 50.78 & 0.547 & 0.187 \\
    \midrule
    \multirow{3}{*}{Qwen2.5-VL-72B-Instruct}
        & Baseline & 66.40 & 39.28 & 0.511 & 0.185 \\
        & Rubric & \textbf{67.19} & \textbf{41.07} & \textbf{0.483} & 0.195 \\
        & CoT & 66.01 & 38.73 & 0.543 & \textbf{0.165} \\
    \bottomrule
  \end{tabular}
  }
\end{table}

\begin{figure*}[t]
    \centering
    \includegraphics[width=\linewidth]{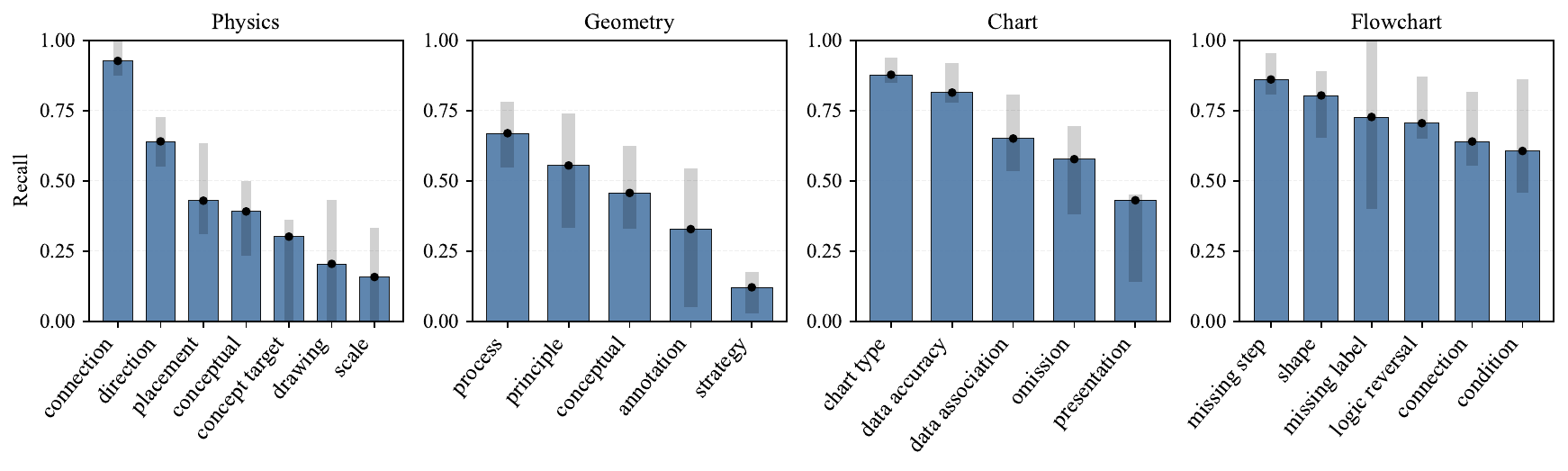}
    \caption{Per-class diagnostic recall across four domains. Bars show per-class recall aggregated over all evaluated models; error bars indicate the interquartile range (25–75\%) of per-model recalls. Category names are shortened for readability; full definitions are provided in Appendix~\ref{app:error_tax}.}
    \label{fig:error_recall}
\end{figure*}

Table~\ref{tab:prompt_main} compares three prompting styles on three representative models: \textbf{Gemini-2.5-Flash}, \textbf{GLM-4.6V}, and \textbf{Qwen2.5-VL-72B-Instruct}. Rubric prompting yields only modest and model-dependent changes: it slightly improves Acc and eb$F_1$ for GLM-4.6V and Qwen2.5-VL-72B-Instruct, but decreases both metrics for Gemini-2.5-Flash. In contrast, CoT prompting consistently hurts performance for all three models. For Gemini-2.5-Flash, CoT reduces Acc from 77.74\% to 76.16\% while increasing FNR from 0.263 to 0.351, indicating more conservative yet less reliable grading.

These results suggest that, in diagram grading, longer reasoning traces do not reliably compensate for noisy visual perception. A plausible explanation is that CoT prompts amplify early misinterpretations of sketch structure or symbols, which then propagate through subsequent reasoning and degrade the final decision. This contrasts with text-only settings where CoT often improves reasoning reliability, and highlights that perceptual uncertainty and cross-modal alignment remain primary bottlenecks for multimodal graders.

\subsubsection{Per-Error-Type Recall Analysis}

We further examine whether MLLMs can diagnose \emph{which} error occurs in an incorrect diagram. Figure~\ref{fig:error_recall} reveals two clear patterns. First, error types tied to salient local cues are consistently easier: connection-related errors in Physics and chart-type mistakes in Charts achieve the highest recall, and Flowchart errors such as missing steps or incorrect shapes are also detected reliably. In contrast, categories that depend on global structure or latent intent remain difficult across domains---including concept-target and scaling errors in Physics, strategy-level mistakes in Geometry, and presentation or omission errors in Charts.

Second, several categories exhibit large variability across models, indicating that fine-grained error diagnosis is not yet a stable capability even among strong MLLMs. Overall, the results highlight that robust diagnostic grading requires deeper structural and semantic understanding beyond surface-level cues.

\section{Conclusion and Outlook}
This work steps beyond the longstanding ``perfect world'' assumption in multimodal evaluation and asks whether MLLMs can act as reliable \textbf{diagram graders} rather than merely \textbf{answer solvers}. With \textbf{SketchJudge}, we instantiate this question across four STEM-related domains, pairing \textbf{messy, hand-drawn student sketches} with \textbf{binary correctness labels} and \textbf{fine-grained, domain-specific error taxonomies}. Our experiments show that state-of-the-art MLLMs still struggle with core grading competencies, including \textbf{robustness to stylistic variation}, \textbf{sensitivity to structural and topological distinctions}, and \textbf{consistent attribution of conceptual errors}. Diagnostic analyses further suggest that current models often behave more like \emph{perceptual pattern recognizers} than \textbf{reasoning-aware graders}.

Looking forward, SketchJudge enables research beyond direct grading. It can serve as a testbed for \textbf{tool-augmented} and \textbf{agentic} multimodal reasoning, where models must plan, externalize intermediate steps, and interact with tools to construct, generate, or verify diagrams. Recent progress on structured reasoning and curriculum-style in-context learning for complex problem solving offers complementary avenues for strengthening such capabilities~\citep{ma-etal-2025-problem}. We hope SketchJudge helps drive progress toward MLLMs capable of \textbf{deeper diagram-level reasoning} and \textbf{feedback-oriented assessment}.

\section*{Broader Impacts}
As a benchmark targeting educational grading scenarios, \textbf{SketchJudge} highlights both the promise and the brittleness of current MLLMs when interacting with \textbf{student-generated content}. Improved diagram-level understanding could support \textbf{scalable formative assessment} and \textbf{personalized feedback}, but high-stakes use requires careful validation and oversight. Models evaluated on \textbf{SketchJudge} should be viewed as \emph{research tools} rather than replacements for human educators. We encourage future work on \textbf{human--AI collaborative workflows}, stronger \textbf{transparency and calibration} in model decisions, and safeguards that ensure equitable and pedagogically sound deployment of multimodal grading technologies.

\section*{Limitations}
SketchJudge relies on a labor-intensive pipeline involving human-written error descriptions, expert-informed taxonomy refinement, and verification of model-assigned labels. While this protocol improves educational plausibility and annotation quality, it also introduces practical constraints on the speed of dataset expansion and the ease of updating taxonomies. In addition, some error categories admit borderline cases where distinctions are subtle and depend on domain judgment, which can introduce residual ambiguity despite careful verification. Finally, SketchJudge focuses on controlled educational-style diagram tasks; extending grading models to broader real-world settings will require further attention to robustness, calibration, and user-facing reliability.

\bibliography{references}
\bibliographystyle{templet}

\appendix

\section{SketchJudge Task Details}
\label{app:task_details}

This appendix supplements the main benchmark description in Section~\ref{sec:sketchjudge_overview} by providing domain-level details for the four task categories in SketchJudge. We summarize (i) the data sources and synthetic problem construction procedure, and (ii) the task characteristics, dataset statistics, and typical error patterns for each domain (Geometry, Physics, Charts, and Flowcharts).


\subsection{Data Sanitization and Content Screening}
\label{app:data_sanitization}

We manually screened all collected samples to remove any personally identifying information (e.g., names) and inappropriate or offensive content. We only retain the diagram content necessary for benchmark construction and evaluation.

\subsection{Data Sources and Synthetic Problem Construction}
\label{app:data_gen}

SketchJudge problems come from two sources: (i) past-exam or textbook-style items collected from the web, and (ii) a constrained synthetic subset limited to \textbf{charts} and \textbf{flowcharts}. The synthetic items were generated with GPT-4o via interactive, multi-turn prompting following domain-specific instructions (rather than a single fixed prompt template). We guided the model to produce diagram-drawing tasks with explicit constraints (e.g., chart type and data values; flowchart steps, decision branches, and standard symbols), and re-generated instances that were ambiguous or violated conventions. Each item is tagged by provenance (web vs.\ generated) to enable source-aware analyses, while main results aggregate across both sources unless otherwise noted. All synthetic problems underwent manual verification to ensure (i) internal consistency of constraints, (ii) adherence to standard diagram conventions, and (iii) solvability under the stated requirements. Instances failing verification were discarded and re-generated rather than edited.

\subsection{Geometry Tasks}

Geometry tasks focus on geometric construction and spatial reasoning under freehand drawing noise (e.g., imprecise angles and ambiguous auxiliary marks). Table~\ref{tab:geo_stats} summarizes the geometry split.

\newcolumntype{C}[1]{>{\centering\arraybackslash}p{#1}}
\newcolumntype{R}[1]{>{\raggedleft\arraybackslash}p{#1}}
\newcolumntype{L}[1]{>{\raggedright\arraybackslash}p{#1}}

\begin{table}[htbp]
    \centering
    \small
    \caption{Geometry domain statistics.}
    \label{tab:geo_stats}
    \begin{tabular}{L{4.5cm}R{2.3cm}}
    \toprule
    \textbf{Statistic} & \textbf{Value} \\
    \midrule
    Problems & 100 \\
    Student answers & 261 \\
    Questions with images & 98 \\
    Reference diagrams & 100 \\
    Avg.\ answers per problem & 2.61 \\
    Synthetic proportion & 0\% \\
    \bottomrule
    \end{tabular}
\end{table}

\paragraph{Typical error patterns.}
Geometry sketches commonly exhibit errors related to \emph{principle violations} (Geometric Principle Error), \emph{missing or incorrect auxiliary constructions} (Operational Process / Strategy Planning Error), and \emph{annotation or representation issues} (Diagram Annotation Error). Many problems also admit multiple valid constructions, making grading harder than direct reference matching.

\subsection{Physics Tasks}

Physics tasks involve circuit diagrams, vector reasoning, and force analysis, requiring both geometric fidelity (e.g., arrow orientation) and semantic correctness (e.g., current flow). Table~\ref{tab:phy_stats} summarizes the physics split.

\begin{table}[htbp]
    \centering
    \small
    \caption{Physics domain statistics.}
    \label{tab:phy_stats}
    \begin{tabular}{L{4.5cm}R{2.3cm}}
    \toprule
    \textbf{Statistic} & \textbf{Value} \\
    \midrule
    Problems & 50 \\
    Student answers & 233 \\
    Questions with images & 49 \\
    Reference diagrams & 50 \\
    Avg.\ answers per problem & 4.66 \\
    Synthetic proportion & 0\% \\
    \bottomrule
    \end{tabular}
\end{table}

\paragraph{Typical error patterns.}
Physics sketches often fail due to incorrect \emph{direction or polarity} (e.g., reversed vectors or circuit polarity), \emph{missing or wrong connections} between elements, and \emph{misplaced components or forces} that break the intended physical setup. Models also struggle with errors that require conceptual interpretation, such as misunderstanding the target quantity or applying an invalid physical principle. In vector-based diagrams, inaccurate relative magnitudes further introduce subtle grading difficulty beyond surface-level appearance.

\subsection{Chart Tasks}

Chart tasks require visualizing numerical data as bar, line, or pie charts. Errors are often visually salient but depend on accurate mapping between values and graphical elements. Table~\ref{tab:chart_stats} summarizes the chart split. 
\textit{(Chart types: bar/line/pie/others = 54/25/19/2.)}

\begin{table}[htbp]
    \centering
    \small
    \caption{Chart domain statistics.}
    \label{tab:chart_stats}
    \begin{tabular}{L{4.5cm}R{2.3cm}}
    \toprule
    \textbf{Statistic} & \textbf{Value} \\
    \midrule
    Problems & 100 \\
    Student answers & 287 \\
    Questions with images & 0 \\
    Reference diagrams & 100 \\
    Avg.\ answers per problem & 2.87 \\
    Synthetic proportion & 15\% \\
    \bottomrule
    \end{tabular}
\end{table}

\paragraph{Typical error patterns.}
Chart sketches typically fail in three ways: choosing an inappropriate chart form for the task, mis-associating categories and values, or encoding the data inaccurately through incorrect scaling and proportions. Even when the overall structure is plausible, charts often omit essential elements such as labels or data entries, or violate basic presentation conventions (e.g., unclear axes or inconsistent annotations), which makes them difficult to grade reliably beyond superficial visual matching.

\subsection{Flowchart Tasks}

Flowchart tasks require understanding procedural structure, control flow, and decision logic. Unlike charts, correctness depends on global topology rather than local appearance. Table~\ref{tab:flow_stats} summarizes the flowchart split. 

\begin{table}[htbp]
    \centering
    \small
    \caption{Flowchart domain statistics.}
    \label{tab:flow_stats}
    \begin{tabular}{L{4.5cm}R{2.3cm}}
    \toprule
    \textbf{Statistic} & \textbf{Value} \\
    \midrule
    Problems & 50 \\
    Student answers & 234 \\
    Questions with images & 0 \\
    Reference diagrams & 50 \\
    Avg.\ answers per problem & 4.68 \\
    Synthetic proportion & 38\% \\
    \bottomrule
    \end{tabular}
\end{table}

\paragraph{Typical error patterns.}
Flowchart answers commonly contain \emph{structural topology issues}, including missing steps, incorrect arrow connections, or reversed decision branches that alter the intended execution logic. In addition, students frequently misuse flowchart symbols (e.g., decision vs.\ process shapes) or omit key labels such as branch outcomes, which makes the logic ambiguous even when most nodes appear correct. These patterns highlight that flowchart grading depends heavily on global control flow rather than local visual similarity.

\section{Error Category Taxonomy }
\label{app:error_tax}

To support fine-grained evaluation, we design a taxonomy of error types covering all four task categories. Tables~\ref{tab:geometry-taxonomy}, \ref{tab:physics-taxonomy}, \ref{tab:flowchart-taxonomy}, and \ref{tab:charts-taxonomy} provides the full list of categories and definitions. This taxonomy serves as the basis for our multi-label annotation protocol described in Section~\ref{sec:annotation}.

\begin{table*}[htbp]
\centering
\small
\begin{tabular}{@{}K{0.24\textwidth} K{0.72\textwidth}@{} K{0.0\textwidth}@{}}
\toprule
\textbf{Error Category} & \textbf{Definition} & \rule{0pt}{2.6ex} \\
\midrule
Conceptual Cognition Error & A misunderstanding of basic geometric concepts or the construction target, leading to drawing an object inconsistent with the task requirements. & \rule{0pt}{6.0ex} \\
\hdashline
Geometric Principle Error & Failure to correctly apply geometric principles or violation of fundamental geometric rules, resulting in a construction that does not meet the requirements. & \rule{0pt}{6.0ex} \\
\hdashline
Operational Process Error & Problems in the use of drawing tools or execution of steps, or the absence of necessary construction traces, leading to a result lacking standardization. & \rule{0pt}{6.0ex} \\
\hdashline
Strategy Planning Error & Lack of a clear construction plan; reasoning steps or the overall approach are flawed, undermining the correctness of the construction. & \rule{0pt}{6.0ex} \\
\hdashline
Diagram Annotation Error & The orientation, connections, annotations, or representation standards of the graph contain errors, leading to disordered relationships among geometric elements and an unclear intent.  & \rule{0pt}{6.0ex}\\
\bottomrule
\end{tabular}
\caption{Error taxonomy in the Geometry category.}
\label{tab:geometry-taxonomy}
\end{table*}

\begin{table*}[htbp]
\centering
\small
\begin{tabular}{@{}K{0.24\textwidth} K{0.72\textwidth}@{} K{0.0\textwidth}@{}}
\toprule
\textbf{Error Category} & \textbf{Definition}  & \rule{0pt}{2.6ex}\\
\midrule
Concept/Target Error & Failure to understand the core objective of the problem, resulting in a diagram that does not match the task requirements. & \rule{0pt}{6.0ex}\\
\hdashline
Direction/Polarity Error & Arrows, directions, or polarity symbols are reversed or inconsistent with physical laws. & \rule{0pt}{6.0ex}\\
\hdashline
Magnitude/Scale Error & The relative magnitudes of vectors or physical quantities are drawn incorrectly, contradicting the problem statement or physical principles. & \rule{0pt}{6.0ex}\\
\hdashline
Connection Error & Incorrect or missing connections between elements, such as forces, components, rays, or nodes, leading to an invalid representation of physical interactions or processes. & \rule{0pt}{6.0ex}\\
\hdashline
Placement Error & Elements, forces, light rays, or components are positioned incorrectly, making the representation inconsistent with the actual physical situation. & \rule{0pt}{6.0ex}\\
\hdashline
Conceptual Misunderstanding & Misapplication of fundamental physical principles leads to a representation that is physically invalid or impossible. & \rule{0pt}{6.0ex}\\
\hdashline
Drawing Convention Error & Violation of standard drawing conventions or established norms in physics education, such as incorrect use of line types or symbols. & \rule{0pt}{6.0ex}\\
\bottomrule
\end{tabular}
\caption{Error taxonomy in the Physics category.}
\label{tab:physics-taxonomy}
\end{table*}

\begin{table*}[htbp]
\centering
\small
\begin{tabular}{@{}K{0.24\textwidth} K{0.72\textwidth}@{} K{0.0\textwidth}@{}}
\toprule
\textbf{Error Category} & \textbf{Definition}  & \rule{0pt}{2.6ex}\\
\midrule
Shape Error & The shape of the flowchart element does not match the standard notation for its function (e.g., using a rectangle instead of a diamond for a decision). & \rule{0pt}{6.0ex}\\
\hdashline
Connection Error & The connectors between elements are drawn incorrectly, such as arrows pointing in the wrong direction or invalid links between nodes. & \rule{0pt}{6.0ex}\\
\hdashline
Missing Step & A required processing step in the flowchart is omitted, leaving the workflow incomplete. & \rule{0pt}{6.0ex}\\
\hdashline
Missing Label & Essential labels are missing or unclear, such as omitting Yes/No labels on decision branches. & \rule{0pt}{6.0ex}\\
\hdashline
Logic Reversal & The logical flow is reversed or misrepresented, such as swapping Yes/No branches or inverting decision outcomes. & \rule{0pt}{6.0ex}\\
\hdashline
Condition Error & The condition itself is incorrect, incomplete, or ambiguously expressed, leading to misinterpretation of the decision point. & \rule{0pt}{6.0ex}\\
\bottomrule
\end{tabular}
\caption{Error taxonomy in the Flowchart category.}
\label{tab:flowchart-taxonomy}
\end{table*}

\begin{table*}[htbp]
\centering
\small
\begin{tabular}{@{}K{0.24\textwidth} K{0.72\textwidth}@{} K{0.0\textwidth}@{}}
\toprule
\textbf{Error Category} & \textbf{Definition}  & \rule{0pt}{2.6ex}\\
\midrule
Inappropriate Chart Type & Selection of an incorrect chart type, inconsistent with the data characteristics or task requirements, leading to confusion in data representation. & \rule{0pt}{6.0ex}\\
\hdashline
Data Association Error & Incorrect organization of data categories, leading to mismatched labels and values. & \rule{0pt}{6.0ex}\\
\hdashline
Data Accuracy Error & Data accuracy is compromised due to encoding or scaling errors during the conversion of data into visual elements. & \rule{0pt}{6.0ex}\\
\hdashline
Omission Error & Omission of essential elements, labels, or data entries, reducing the completeness of the chart. & \rule{0pt}{6.0ex}\\
\hdashline
Presentation Convention Error & Failure to follow labeling or standardization conventions, causing unclear or inaccurate chart presentation. & \rule{0pt}{6.0ex}\\
\bottomrule
\end{tabular}
\caption{Error taxonomy in the Charts category.}
\label{tab:charts-taxonomy}
\end{table*}

\begin{table}[t]
    \centering
    \small
    \caption{Mapping between full error type names and shortened labels used in Figure~\ref{fig:error_recall}.}
    \label{tab:alias_mapping}
    \resizebox{\columnwidth}{!}{
    \begin{tabular}{lll}
        \toprule
        \textbf{Domain} & \textbf{Full Name} & \textbf{Alias} \\
        \midrule
        Physics & Connection Error & connection \\
        Physics & Direction Polarity Error & direction \\
        Physics & Placement Error & placement \\
        Physics & Conceptual Misunderstanding & conceptual \\
        Physics & Concept Target Error & concept target \\
        Physics & Drawing Convention Error & drawing \\
        Physics & Magnitude Scale Error & scale \\
        \midrule
        Geometry & Operational Process Error & process \\
        Geometry & Geometric Principle Error & principle \\
        Geometry & Conceptual Cognition Error & conceptual \\
        Geometry & Diagram Annotation Error & annotation \\
        Geometry & Strategy Planning Error & strategy \\
        \midrule
        Chart & Inappropriate Chart Type & chart type \\
        Chart & Data Accuracy Error & data accuracy \\
        Chart & Data Association Error & data association \\
        Chart & Omission Error & omission \\
        Chart & Presentation Convention Error & presentation \\
        \midrule
        Flowchart & Missing Step & missing step \\
        Flowchart & Shape Error & shape \\
        Flowchart & Missing Label & missing label \\
        Flowchart & Logic Reversal & logic reversal \\
        Flowchart & Connection Error & connection \\
        Flowchart & Condition Error & condition \\
        \bottomrule
    \end{tabular}
    }
\end{table}

\section{Prompt Templates}
\label{app:prompts}

This appendix documents all prompt templates used in SketchJudge, covering both the construction of annotation taxonomies and the evaluation of multimodal models. In particular, we provide the prompts used for (i) inducing domain-specific error taxonomies and assigning error labels during annotation, and (ii) evaluating model grading performance under different input settings.

\subsection{Prompts for Taxonomy Induction and Annotation}
\label{app:annotation_prompts}

We employ prompt-based interaction with GPT-4o to support both stages of the annotation protocol described in Section~\ref{sec:annotation}. Specifically, GPT-4o is used to (i) cluster annotator-written error descriptions into candidate error categories during bottom-up taxonomy induction, and (ii) map each error description to one or more error types under the finalized taxonomy. In both cases, the model operates solely on textual error descriptions and has no access to the corresponding sketches.

The full prompt templates used for taxonomy induction and taxonomy-based labeling are listed in Tables~\ref{tab:prompt_taxonomy_induction} and~\ref{tab:prompt_taxonomy_labeling}.

\begin{table}[hbtp]
\centering
\small
\setlength{\tabcolsep}{6pt}
\caption{Prompt used for draft taxonomy induction from annotator-written error descriptions.}
\label{tab:prompt_taxonomy_induction}
\begin{tabular}{p{0.18\linewidth} p{0.7\linewidth}}
\toprule
\textbf{Goal} & \textbf{Draft taxonomy induction} \\
\midrule
\textbf{Prompt} &
\texttt{
We are building an error taxonomy for grading hand-drawn diagrams in \{\{DOMAIN\}\}.
Below are error descriptions written by annotators for incorrect student sketches.
Please propose a concise set of error categories that can cover these mistakes.
For each category, give a short name and its corresponding description.
} \\
\midrule
\textbf{Input} &
\texttt{[[ERROR\_DESCRIPTIONS\_LIST]]} \\
\bottomrule
\end{tabular}
\end{table}

\begin{table}[hbtp]
\centering
\small
\setlength{\tabcolsep}{6pt}
\caption{Prompt used for taxonomy-based labeling from text-only error descriptions (diagram withheld).}
\label{tab:prompt_taxonomy_labeling}
\begin{tabular}{p{0.18\linewidth} p{0.7\linewidth}}
\toprule
\textbf{Goal} & \textbf{Assign taxonomy labels} \\
\midrule
\textbf{Prompt} &
\texttt{
You are given an error taxonomy and one free-form error description written by an annotator. Please assign exactly one taxonomy label that best matches the description based only on the text. Use only labels from the given taxonomy. Return the label only (no explanation).
} \\
\midrule
\textbf{Input} &
\texttt{[[TAXONOMY\_BULLETED]]\newline
[[ERROR\_DESCRIPTIONS\_LIST]]} \\
\bottomrule
\end{tabular}
\end{table}

\subsection{Prompts for Model Evaluation}
\label{app:evaluation_prompts}

We next describe the prompt templates used to evaluate multimodal models on SketchJudge. These prompts define the grading task, available inputs, and required structured outputs under different task settings.

\subsection{Task Settings}

We consider four canonical input configurations:

\begin{table}[hbtp]
\centering
\small
\caption{Four canonical input configurations for SketchJudge grading.}
\label{tab:task_settings}
    \begin{tabular}{L{0.8cm}C{2.5cm}C{2.9cm}}
    \toprule
    \textbf{Case} & \textbf{Question Content} & \textbf{Reference Diagram} \\
    \midrule
    1 & Text-only & Not provided \\
    2 & Text-only & Provided \\
    3 & Contains an image & Not provided \\
    4 & Contains an image & Provided \\
    \bottomrule
    \end{tabular}
\end{table}

For each case, the Baseline prompt provides the minimal necessary  grading instructions. The full Baseline templates for all four cases appear in Figures~\ref{fig:prompt_case1}--\ref{fig:prompt_case4}. Rubric-based and Chain-of-Thought prompts extend the Baseline by adding explicit evaluation rubrics or stepwise reasoning guidance while preserving the same input structure and JSON output specification. These templates are listed in Tables~\ref{tab:prompt_rubric} and~\ref{tab:prompt_step}.

\subsection{Baseline Prompt Templates}

\begin{figure*}[htbp]
  \centering
  \includegraphics[width=0.82\linewidth]{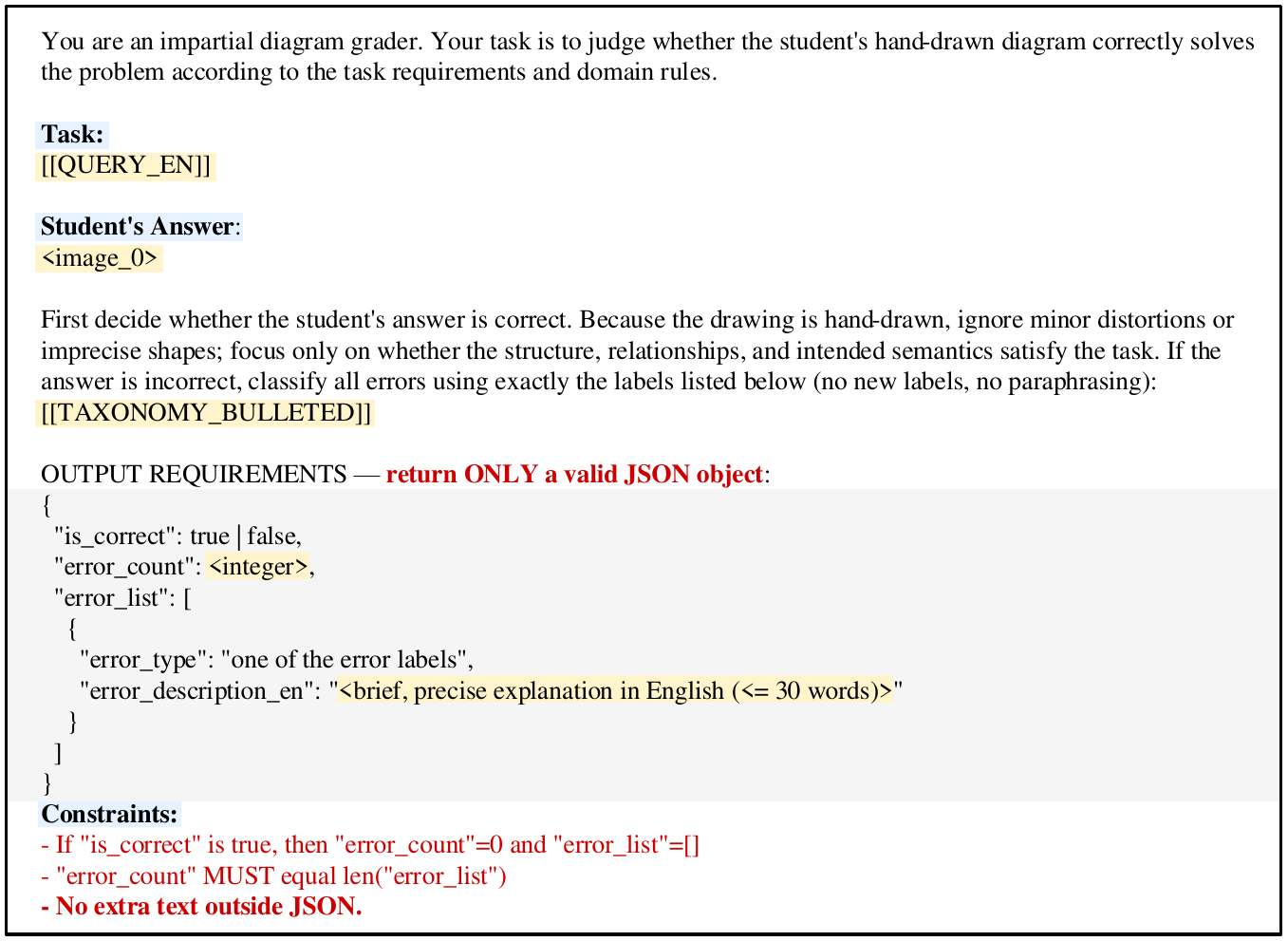}
  \caption{Prompt template for Case 1 (Text-only, No Reference).}
  \label{fig:prompt_case1}
\end{figure*}

\begin{figure*}[htbp]
  \centering
  \includegraphics[width=0.82\linewidth]{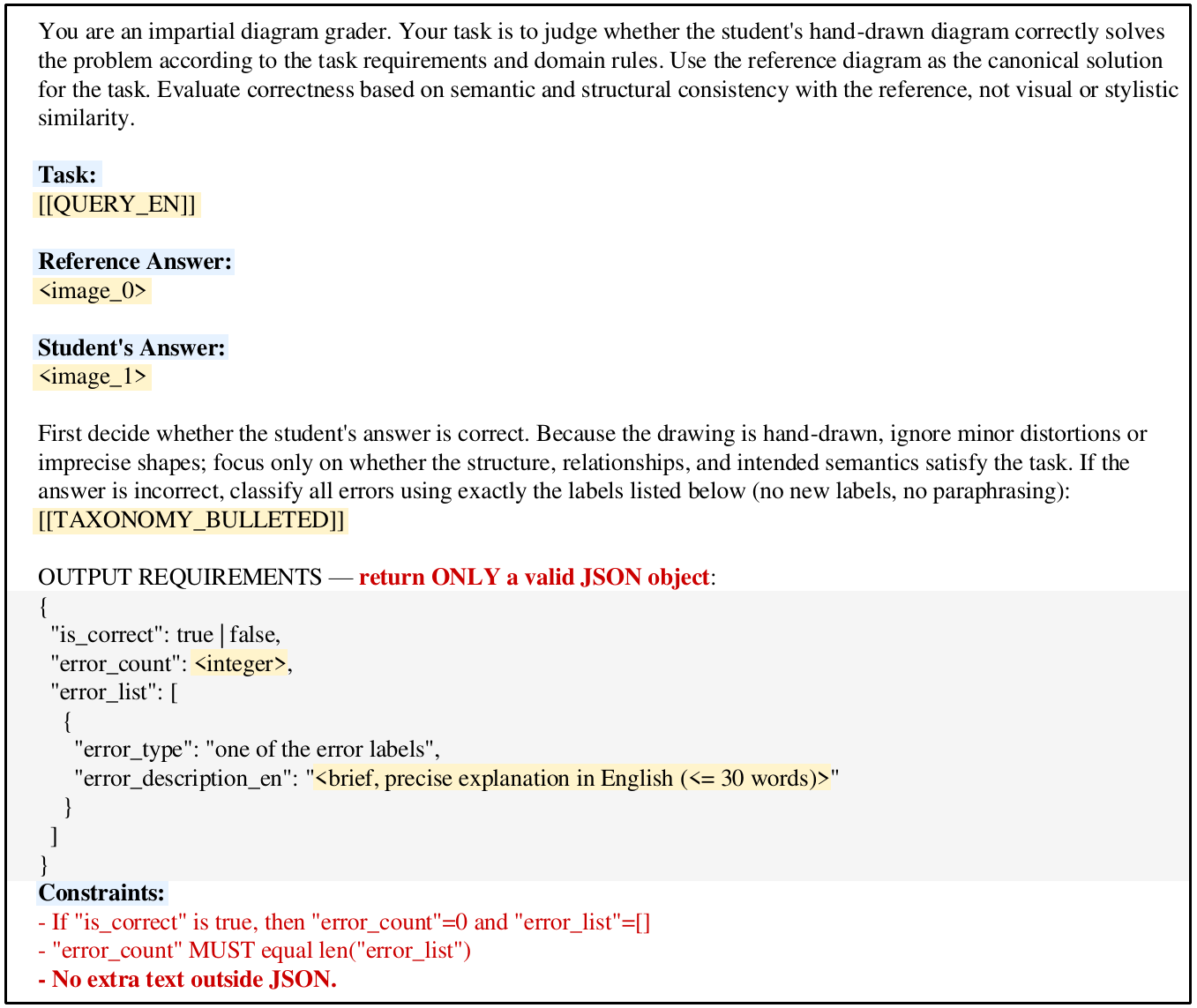}
  \caption{Prompt template for Case 2 (Text-only, With Reference).}
  \label{fig:prompt_case2}
\end{figure*}

\begin{figure*}[htbp]
  \centering
  \includegraphics[width=0.82\linewidth]{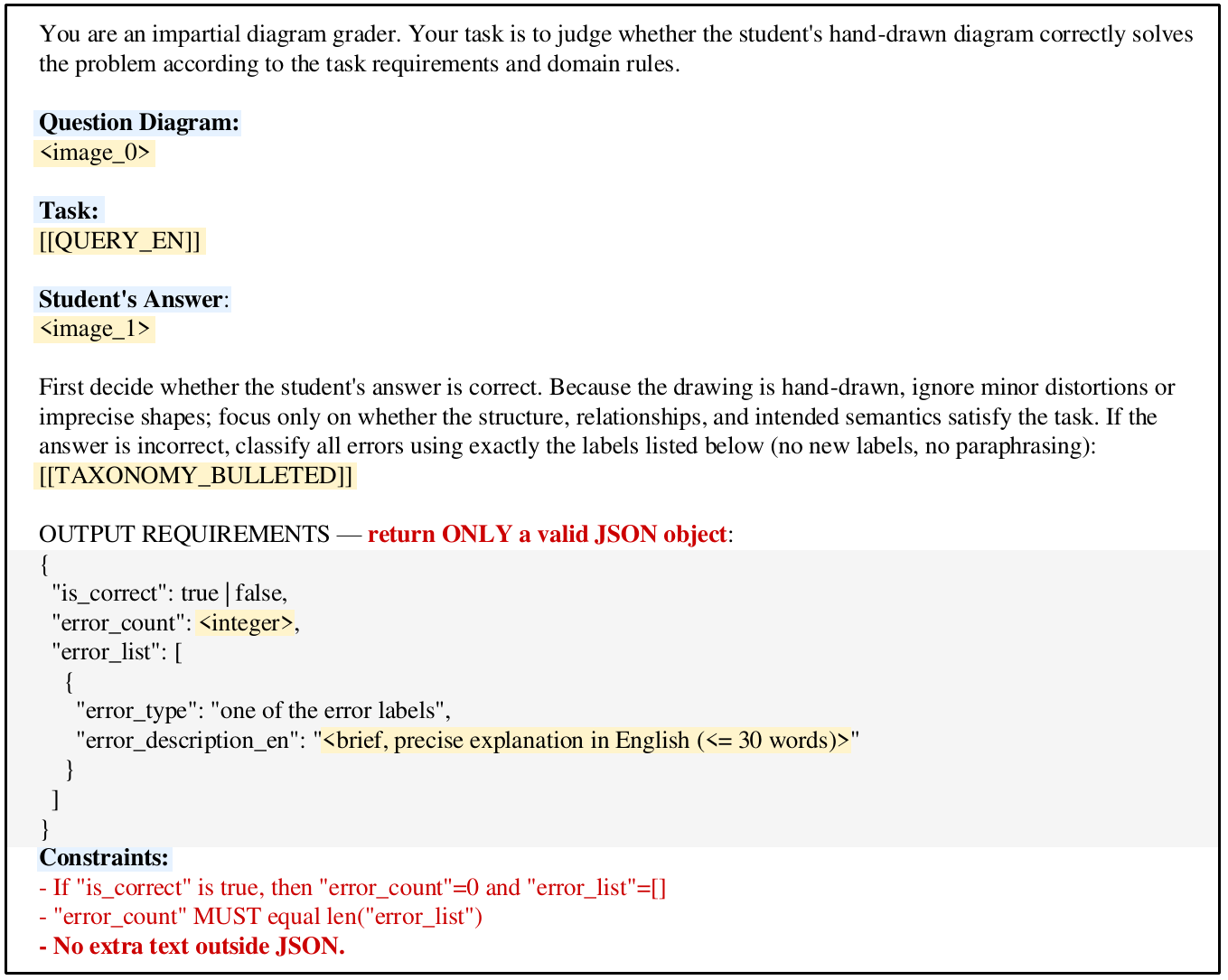}
  \caption{Prompt template for Case 3 (Diagram, No Reference).}
  \label{fig:prompt_case3}
\end{figure*}

\begin{figure*}[htbp]
  \centering
  \includegraphics[width=0.82\linewidth]{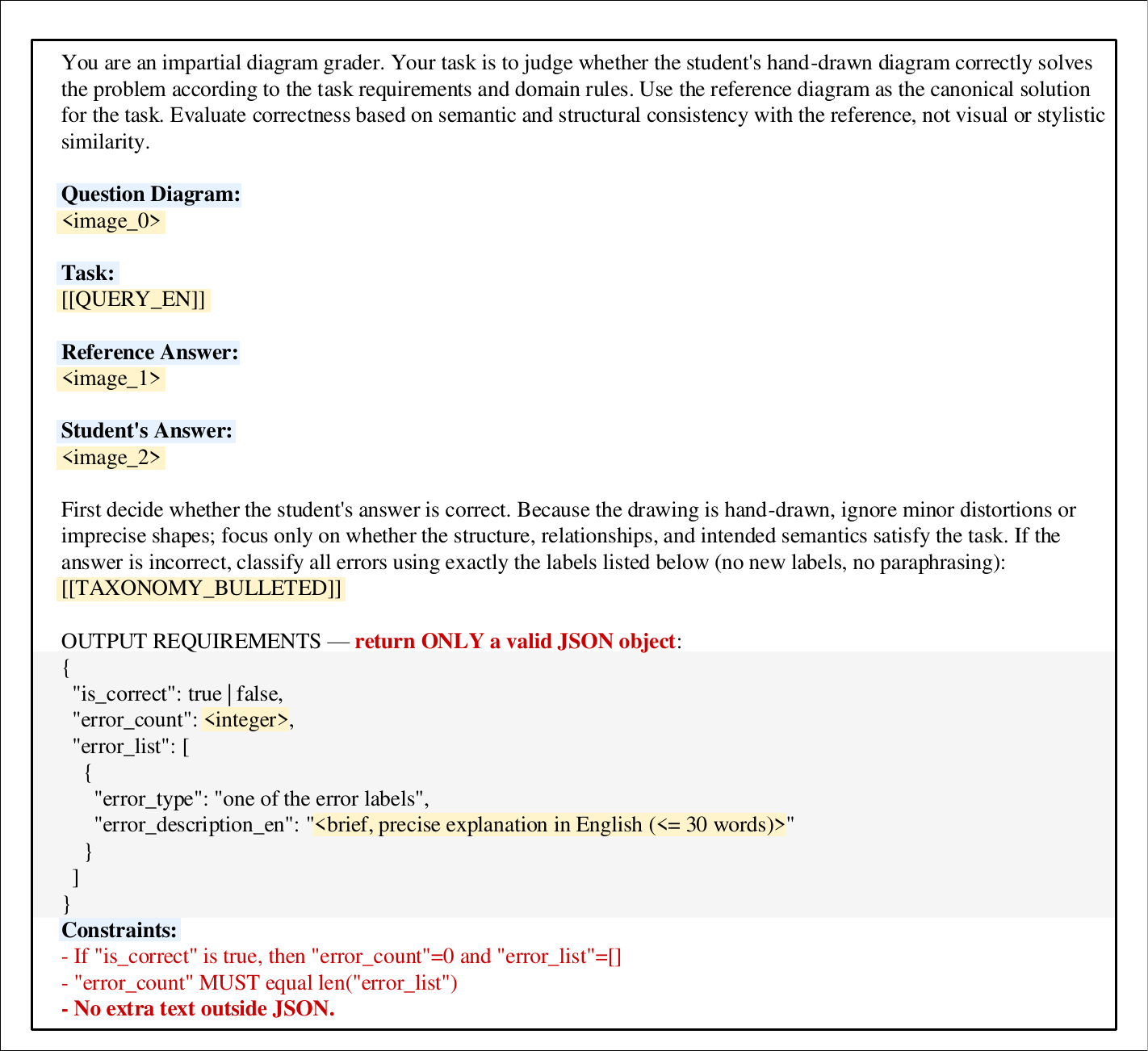}
  \caption{Prompt template for Case 4 (Diagram, With Reference).}
  \label{fig:prompt_case4}
\end{figure*}

Figures~\ref{fig:prompt_case1}--\ref{fig:prompt_case4} show the complete Baseline prompts used in our main experiments. They define the core task setting, model role, available images, and the structured JSON output expected from all models.

\subsection{Rubric-Based Prompt Templates}

Table~\ref{tab:prompt_rubric} provides the full Rubric-based variants. These prompts supply explicit grading criteria describing which aspects of the diagram should be checked. Aside from the additional rubric, all other components—including the task description, image inputs, and JSON output format—remain identical to the Baseline family.

\begin{table*}[htbp]
\centering
\small
\caption{Rubric-based prompt templates. We share a common rubric prompt and vary only the input configuration. T: text-only question; D: question includes a diagram; R: reference diagram provided; N: no reference.}
\label{tab:prompt_rubric}
\setlength{\tabcolsep}{6pt}
\begin{tabular}{cclK{0.0\textwidth}}
\toprule
\textbf{Case} & \textbf{Setting} & \textbf{Inputs inserted into the shared rubric prompt} \\
\midrule
1 & (T, N) & \textbf{Task}: [[QUERY\_EN]]; \textbf{Student}: \texttt{<image\_0>} & \rule{0pt}{3ex} \\
2 & (T, R) & \textbf{Task}: [[QUERY\_EN]]; \textbf{Ref}: \texttt{<image\_0>}; \textbf{Student}: \texttt{<image\_1>} & \rule{0pt}{3ex}\\
3 & (D, N) & \textbf{Q-diagram}: \texttt{<image\_0>}; \textbf{Task}: [[QUERY\_EN]]; \textbf{Student}: \texttt{<image\_1>} & \rule{0pt}{3ex} \\
4 & (D, R) & \textbf{Q-diagram}: \texttt{<image\_0>}; \textbf{Task}: [[QUERY\_EN]]; \textbf{Ref}: \texttt{<image\_1>}; \textbf{Student}: \texttt{<image\_2>} & \rule{0pt}{3ex}\\
\bottomrule
\end{tabular}

\vspace{4pt}
\begin{minipage}{0.98\linewidth}
\noindent
\textbf{Shared rubric prompt (\textit{NoRef}).}
\begin{quote}
\footnotesize\ttfamily
You are an impartial diagram grader. Follow the rubric below to determine whether the student's hand-drawn diagram correctly solves the task.\newline
RUBRIC -- A correct diagram MUST:\newline
(1) include all required task elements;\newline
(2) preserve correct spatial, structural, and logical relationships;\newline
(3) convey domain-appropriate semantics;\newline
(4) be complete and coherent, even with typical hand-drawn imperfections.\newline
A diagram is incorrect if it violates any required structural, semantic, or logical condition.\newline
[Insert case-specific inputs here]\newline
Evaluate correctness using the rubric. Ignore normal hand-drawn inaccuracies.
If incorrect, classify ALL errors using exactly the labels below (no new labels, no paraphrasing):\newline
[[TAXONOMY\_BULLETED]].\newline
(Return ONLY the JSON object specified in the baseline template.)
\end{quote}

\noindent
\textbf{Shared rubric prompt (\textit{WithRef}).}
\begin{quote}
\footnotesize\ttfamily
You are an impartial diagram grader. Follow the rubric below to determine whether the student's hand-drawn diagram correctly solves the task. Use the reference diagram as the canonical solution, and evaluate semantic and structural consistency rather than visual similarity.\newline
RUBRIC -- A correct diagram MUST:\newline
(1) include all required task elements;\newline
(2) preserve correct spatial, structural, and logical relationships;\newline
(3) convey domain-appropriate semantics;\newline
(4) be complete and coherent, even with typical hand-drawn imperfections.\newline
A diagram is incorrect if it violates any required structural, semantic, or logical condition.\newline
[Insert case-specific inputs here]\newline
Evaluate correctness using the rubric. Ignore normal hand-drawn inaccuracies.
If incorrect, classify ALL errors using exactly the labels below (no new labels, no paraphrasing):\newline
[[TAXONOMY\_BULLETED]].\newline
(Return ONLY the JSON object specified in the baseline template.)
\end{quote}
\end{minipage}
\end{table*}

\begin{table*}[htbp]
\centering
\small
\caption{Chain-of-Thought reasoning templates. All variants share the same JSON output format as the baseline family. T: Text-only, D: Diagram in question, R: With reference, N: No reference.}
\label{tab:prompt_step}
\setlength{\tabcolsep}{6pt}
\begin{tabular}{cclK{0.0\textwidth}}
\toprule
\textbf{Case} & \textbf{Setting} & \textbf{Inputs inserted into the shared rubric prompt} \\
\midrule
1 & (T, N) & \textbf{Task}: [[QUERY\_EN]]; \textbf{Student}: \texttt{<image\_0>} & \rule{0pt}{3ex} \\
2 & (T, R) & \textbf{Task}: [[QUERY\_EN]]; \textbf{Ref}: \texttt{<image\_0>}; \textbf{Student}: \texttt{<image\_1>} & \rule{0pt}{3ex}\\
3 & (D, N) & \textbf{Q-diagram}: \texttt{<image\_0>}; \textbf{Task}: [[QUERY\_EN]]; \textbf{Student}: \texttt{<image\_1>} & \rule{0pt}{3ex} \\
4 & (D, R) & \textbf{Q-diagram}: \texttt{<image\_0>}; \textbf{Task}: [[QUERY\_EN]]; \textbf{Ref}: \texttt{<image\_1>}; \textbf{Student}: \texttt{<image\_2>} & \rule{0pt}{3ex}\\
\bottomrule
\end{tabular}

\vspace{4pt}
\begin{minipage}{0.98\linewidth}
\noindent
\textbf{Shared rubric prompt (\textit{NoRef}).}
\begin{quote}
\footnotesize\ttfamily
You are an impartial diagram grader. Carefully evaluate the student's hand-drawn diagram using step-by-step internal reasoning before producing the final JSON output.\newline
INTERNAL REASONING INSTRUCTIONS (do not reveal):\newline
- First interpret the task requirements and the reference diagram.\newline
- Then analyze the student's diagram step by step:\newline
1. Identify all required elements.\newline
2. Check structural and spatial relationships.\newline
3. Evaluate domain-specific correctness (geometry, physics, charts, flowcharts, etc.).\newline
4. Determine semantic equivalence to the reference (not stylistic similarity) and identify any conceptual or procedural flaws.\newline
- Decide correctness only after completing all reasoning steps. Ignore normal hand-drawn inaccuracies.\newline
- Do NOT reveal the reasoning; output only the final JSON result.\newline
[Insert case-specific inputs here]\newline
If the answer is incorrect, classify ALL errors using exactly the labels listed below (no new labels, no paraphrasing):\newline
[[TAXONOMY\_BULLETED]].\newline
(Return ONLY the JSON object specified in the baseline template.)
\end{quote}

\noindent
\textbf{Shared rubric prompt (\textit{WithRef}).}
\begin{quote}
\footnotesize\ttfamily
You are an impartial diagram grader. Carefully evaluate the student's hand-drawn diagram using step-by-step internal reasoning before producing the final JSON output. Use the reference diagram as the canonical solution.\newline
INTERNAL REASONING INSTRUCTIONS (do not reveal):\newline
- First interpret the task requirements and the reference diagram.\newline
- Then analyze the student's diagram step by step:\newline
1. Identify all required elements.\newline
2. Check structural and spatial relationships.\newline
3. Evaluate domain-specific correctness (geometry, physics, charts, flowcharts, etc.).\newline
4. Determine semantic equivalence to the reference (not stylistic similarity) and identify any conceptual or procedural flaws.\newline
- Decide correctness only after completing all reasoning steps. Ignore normal hand-drawn inaccuracies.\newline
- Do NOT reveal the reasoning; output only the final JSON result.\newline
[Insert case-specific inputs here]\newline
If the answer is incorrect, classify ALL errors using exactly the labels listed below (no new labels, no paraphrasing):\newline
[[TAXONOMY\_BULLETED]].\newline
(Return ONLY the JSON object specified in the baseline template.)
\end{quote}
\end{minipage}
\end{table*}

\subsection{Chain-of-Thougth Prompt Templates}

Table~\ref{tab:prompt_step} lists the Chain-of-Thought prompts, which require the model to explicitly articulate a multi-stage reasoning process before producing the final JSON output. This design isolates the effect of chain-of-thought–style guidance while preserving identical output constraints.

\begin{table*}[t]
\centering
\small
\setlength{\tabcolsep}{6pt}
\caption{Evaluated multimodal large language models (MLLMs). ``Provider'' denotes the organization releasing the model. Open/Closed indicates whether model weights are publicly available.}
\label{tab:models}
\resizebox{2.1\columnwidth}{!}{
\begin{tabular}{@{}c c c c c K{0.5\linewidth} K{0.0\textwidth}@{}}
\toprule
\textbf{ID} & \textbf{Model} & \textbf{Provider} & \textbf{Open} & \textbf{Size} & \textbf{Link} \\
\midrule
1  & Claude-3.7-Sonnet & Anthropic & Closed & -- & \url{https://www.anthropic.com/claude} & \rule{0pt}{5ex} \\
2  & Gemini-2.5-Flash & Google & Closed & -- & \url{https://ai.google.dev/gemini-api/docs/models} & \rule{0pt}{5ex} \\
3  & GPT-5 & OpenAI & Closed & -- & \url{https://platform.openai.com/docs/models} & \rule{0pt}{5ex} \\
4  & GLM-4.6V & Zhipu AI & Open & -- & \url{https://glm-v.com/} & \rule{0pt}{5ex} \\
5  & Qwen2.5-VL-72B-Instruct & Qwen & Open & 72B & \url{https://huggingface.co/Qwen/Qwen2.5-VL-72B-Instruct} & \rule{0pt}{5ex} \\
6  & Qwen2.5-VL-7B-Instruct & Qwen & Open & 7B & \url{https://huggingface.co/Qwen/Qwen2.5-VL-7B-Instruct} & \rule{0pt}{5ex} \\
7  & Qwen2.5-VL-3B-Instruct & Qwen & Open & 3B & \url{https://huggingface.co/Qwen/Qwen2.5-VL-3B-Instruct} & \rule{0pt}{5ex} \\
8  & Gemma-3-27B-it & Google & Open & 27B & \url{https://huggingface.co/google/gemma-3-27b-it} & \rule{0pt}{5ex} \\
9  & Gemma-3-4B-it & Google & Open & 4B & \url{https://huggingface.co/google/gemma-3-4b-it} & \rule{0pt}{5ex} \\
10 & Llama-4-Scout-17B-16E-Instruct & Meta & Open & 17B & \url{https://huggingface.co/meta-llama/Llama-4-Scout-17B-16E-Instruct} & \rule{0pt}{5ex} \\
11 & Doubao-Seed-1.6-Vision & ByteDance & Closed & -- & \url{https://www.volcengine.com/product/doubao} & \rule{0pt}{5ex} \\
12 & ERNIE-4.5-Turbo-VL & Baidu & Closed & -- & \url{https://cloud.baidu.com/product/wenxinworkshop} & \rule{0pt}{5ex} \\
13 & Mistral-Large-3 & Mistral AI & Closed & -- & \url{https://docs.mistral.ai/getting-started/models} & \rule{0pt}{5ex} \\
14 & o3 & OpenAI & Closed & -- & \url{https://platform.openai.com/docs/models} & \rule{0pt}{5ex} \\
15 & Qianfan-VL-70B & Baidu & Closed & 70B & \url{https://cloud.baidu.com/product/qianfan} & \rule{0pt}{5ex} \\
16 & Qianfan-VL-8B & Baidu & Closed & 8B & \url{https://cloud.baidu.com/product/qianfan} & \rule{0pt}{5ex} \\
\bottomrule
\end{tabular}
}
\end{table*}

\section{Evaluated Models}
\label{app:models}

Table~\ref{tab:models} lists all evaluated MLLMs, spanning both open-weight and closed-source systems across a range of scales. All models are tested in the same zero-shot setup with identical prompts and deterministic decoding (temperature~=~0). Model IDs are used for compact references in figures and diagnostic analyses.

\paragraph{Output Robustness and Format Compliance}
During evaluation, all models were instructed to return a strict JSON-only output. In practice, a small number of responses from several models occasionally included extra text (e.g., brief reasoning traces or wrapper sentences) outside the requested JSON block. These formatting deviations were rare relative to the full test set, and the underlying JSON content was typically still well-formed and semantically consistent with the intended prediction.

To ensure reliable scoring, we applied a deterministic post-processing step that extracts and validates the final JSON object, and we manually inspected the few remaining failures to correct obvious formatting issues. We report results based on the recovered JSON outputs. Overall, this issue had negligible impact on model performance, but it highlights that robust instruction-following and structured output remain non-trivial even for strong MLLMs.

\section{Metric Definitions}
\label{app:metrics}

We provide complete definitions of the evaluation metrics used in our study. Binary grading metrics capture overall correctness and class balance, while multi-label metrics assess fine-grained error-type recognition.

\subsection{Binary Grading Metrics}
We cast grading as a two-class classification problem with 
$y=1$ denoting gold \emph{Correct} and $y=0$ denoting gold \emph{Incorrect}. For $N$ submissions and predictions $\hat y_i$, we define:

\paragraph{Confusion counts.}
\begin{itemize}
    \item True Positive (TP): \#\{ $y{=}1,\ \hat y{=}1$ \}
    \item False Positive (FP): \#\{ $y{=}0,\ \hat y{=}1$ \}
    \item True Negative (TN): \#\{ $y{=}0,\ \hat y{=}0$ \}
    \item False Negative (FN): \#\{ $y{=}1,\ \hat y{=}0$ \}
\end{itemize}

\paragraph{Accuracy.}
Overall proportion of correct decisions:
\begin{equation}
    \mathrm{Accuracy} = \frac{TP+TN}{TP+FP+TN+FN}.
\end{equation}

\paragraph{MacroF1$_{\text{bin}}$.}
To balance both classes, we compute an F1 score for each class and average:
\begin{equation}
    \mathrm{F1}_{pos} = \frac{2 \cdot TP}{2 \cdot TP + FP + FN},
\end{equation}
\begin{equation}
    \mathrm{F1}_{neg} = \frac{2 \cdot TN}{2 \cdot TN + FP + FN},
\end{equation}
\begin{equation}
    \mathrm{MacroF1}_{\text{bin}} = \frac{1}{2}\big(\mathrm{F1}_{pos} + \mathrm{F1}_{neg}\big).
\end{equation}
Here, $\mathrm{F1}_{pos}$ measures recognition of gold-correct answers, 
while $\mathrm{F1}_{neg}$ measures recognition of gold-incorrect answers.

\paragraph{Error tendencies.}
To analyze systematic biases, we report class-specific error rates in terms of false negatives and false positives. The false negative rate (FNR) reflects a tendency toward over-strictness (i.e., failing to recognize correct answers), while the false positive rate (FPR) captures over-leniency (i.e., accepting incorrect answers).

\begin{equation}
    \mathrm{FNR} = \frac{\mathrm{FN}}{\mathrm{FN} + \mathrm{TP}},
\end{equation}

\begin{equation}
    \mathrm{FPR} = \frac{\mathrm{FP}}{\mathrm{FP} + \mathrm{TN}}.
\end{equation}

\paragraph{Matthews Correlation Coefficient (MCC).}
We also compute the Matthews Correlation Coefficient as follows:

\begin{equation}
\mathrm{MCC} =
\begin{aligned}[t]
  &\frac{TP\!\cdot\!TN - FP\!\cdot\!FN}{\sqrt{(TP+FP)(TP+FN)}}\\[4pt]
  &\times \frac{1}{\sqrt{(TN+FP)(TN+FN)}}.
\end{aligned}
\end{equation}

MCC ranges from $-1$ (inverse prediction) to $1$ (perfect), with $0$ indicating random performance.

\subsection{Error-Type Recognition Metrics}
For submissions judged incorrect, models must also predict specific error categories.
Each subject domain (e.g., physics, geometry) defines its own taxonomy, which we merge into a \emph{namespaced} global label set $C_{\text{global}}$ (e.g., \texttt{physics::omission\_error}) to ensure uniqueness across domains.

Let $y_{i,c}, \hat y_{i,c} \in \{0,1\}$ denote the gold and predicted labels for each $c \in C_{\text{global}}$. We use $TP_c$, $FP_c$, and $FN_c$ to denote counts aggregated over all instances for class $c$.

\paragraph{MacroF1$_{\text{err}}$.}
We report Macro-F1 over all error labels with at least one positive gold instance:
\begin{equation}
\begin{aligned}[t]
    &\text{Macro}F1_{err} = \frac{1}{|C^+|} \sum_{c \in C^+} F1_c, \\[4pt]
    &\quad C^+ = \{c : TP_c + FN_c > 0\}
\end{aligned}
\end{equation}

\paragraph{MicroF1$_{\text{err}}$.}
This variant aggregates counts across all labels before computing F1:
\begin{equation}
    \text{Micro}F1_{err} \!=\! \frac{2 \cdot \sum_c TP_c}{2 \cdot \sum_c TP_c \! + \! \sum_c FP_c \!+ \! \sum_c FN_c}.
\end{equation}

\paragraph{Example-based F1 (eb$F_1$).}
To evaluate the overall error detection capability for each instance, we compute the example-based F1 score. This metric measures the overlap between the predicted error set and the gold error set for each instance and averages over all instances. Given a gold set $G_i$ and the corresponding predicted set $P_i$ for instance $i$ (both masked by applicability), we define the example-based F1 as:

\begin{equation}
F1_i =
\begin{cases}
1.0, & \text{if } |G_i| = |P_i| = 0, \\
\frac{2 \cdot |G_i \cap P_i|}{|G_i| + |P_i|}, & \text{otherwise}
\end{cases}
\end{equation}where $G_i$ and $P_i$ are the sets of true and predicted errors for instance $i$. The overall example-based F1 score is then averaged across all instances:

\begin{equation}
\text{eb}F_1 = \frac{1}{N} \sum_{i=1}^{N} F1_i,
\end{equation}
where $N$ is the total number of instances. $N$ captures the completeness of the error set prediction for each instance, making it especially relevant for tasks with a list of error types per instance.

\paragraph{Per-class diagnostic recall.}
To analyze which conceptual errors models are able to identify, we further report per-class recall for each error category. For an error class $c$, recall is defined as the ratio of correctly recovered instances to the total number of gold instances of $c$, aggregated across all evaluated models:
\begin{equation}
\mathrm{Recall}_c =
\frac{\sum_m TP_{m,c}}{\sum_m (TP_{m,c} + FN_{m,c})},
\end{equation}
where $TP_{m,c}$ and $FN_{m,c}$ denote the true positives and false negatives for class $c$ produced by model $m$, computed only on submissions incorrect in both gold annotations and model predictions.

In addition to the aggregated recall, we report cross-model variability using the interquartile range (25--75\%) of per-model recalls for each class. This statistic reflects how consistently different models identify a given error type, independent of class frequency.







\section{Full Experimental Results}
\label{app:full_results}

This appendix provides extended experimental results omitted from the main text for space reasons. In particular, we report (i) complete leaderboards for all evaluated models across all metrics and domains, (ii) full ablation results for key evaluation settings (e.g., with vs.\ without reference answers), and (iii) additional qualitative and diagnostic analyses that complement the main findings. These results are intended to improve transparency, facilitate reproducibility, and enable deeper inspection of model behaviors beyond the headline numbers reported in the main paper.

\subsection{Full Leaderboard (All Models × All Metrics)}
Tables~\ref{tab:mcc_f1b_withref}--\ref{tab:maf1_mif1_withref} report the complete leaderboard under the \textit{WithRef} setting across all evaluated models and domains. These tables complement the main leaderboard in Table~\ref{tab:main_results} by providing a fuller view of grading behavior beyond accuracy.

\begin{table*}[hbtp]
  \centering
  \caption{MCC (range $[-1,1]$) and Macro$F1_{bin}$ ($\times100$) for all evaluated models under the \textit{WithRef} setting. \textbf{Bold} and \underline{underline} indicate the best and second-best scores among models.}
  \label{tab:mcc_f1b_withref}
  \adjustbox{max width=\textwidth}{
  \begin{tabular}{lcccccccccc}
    \toprule
    \multirow{2}{*}{\textbf{Model}} & \multicolumn{2}{c}{\textbf{Physics}} & \multicolumn{2}{c}{\textbf{Geometry}} & \multicolumn{2}{c}{\textbf{Chart}} & \multicolumn{2}{c}{\textbf{Flowchart}} & \multicolumn{2}{c}{\textbf{overall}}\\
    \cmidrule(lr){2-3}\cmidrule(lr){4-5}\cmidrule(lr){6-7}\cmidrule(lr){8-9}\cmidrule(lr){10-11}
    & MCC $\uparrow$ & Macro$F1_{bin}$ $\uparrow$
    & MCC $\uparrow$ & Macro$F1_{bin}$ $\uparrow$
    & MCC $\uparrow$ & Macro$F1_{bin}$ $\uparrow$
    & MCC $\uparrow$ & Macro$F1_{bin}$ $\uparrow$
    & MCC $\uparrow$ & Macro$F1_{bin}$ $\uparrow$ \\
    \midrule
    \multicolumn{11}{c}{\textit{\cellcolor{gray!10}Open-source models}} \\
    Llama-4-Scout-17B-16E-Instruct & 0.284 & 63.48 & 0.375 & 68.56 & 0.451 & 70.69 & 0.472 & 73.41 & 0.390 & 69.16 \\
    Gemma-3-4B-it & 0.169 & 53.12 & 0.146 & 44.44 & 0.319 & 50.17 & 0.065 & 40.40 & 0.181 & 47.66 \\
    Gemma-3-27B-it & 0.225 & 56.83 & 0.318 & 65.10 & 0.408 & 61.07 & 0.409 & 66.99 & 0.333 & 62.76 \\
    Qwen2.5-VL-3B-Instruct & 0.225 & 56.83 & 0.318 & 65.10 & 0.408 & 61.07 & 0.409 & 66.99 & 0.333 & 62.76 \\
    Qwen2.5-VL-7B-Instruct & 0.244 & 53.35 & 0.045 & 36.71 & 0.297 & 64.41 & 0.378 & 68.78 & 0.232 & 59.90 \\
    Qwen2.5-VL-72B-Instruct & 0.163 & 48.67 & 0.176 & 51.27 & 0.544 & 74.84 & 0.462 & 72.31 & 0.323 & 64.84 \\
    Qianfan-VL-8B & -0.038 & 36.68 & 0.095 & 50.90 & 0.482 & 69.58 & 0.371 & 63.57 & 0.258 & 56.82 \\
    Qianfan-VL-70B & 0.139 & 54.51 & 0.209 & 60.42 & 0.441 & 66.95 & 0.493 & 73.00 & 0.304 & 64.89 \\
    GLM-4.6V & 0.207 & 59.36 & 0.264 & 54.91 & 0.512 & 75.57 & 0.568 & 76.23 & 0.384 & 67.55 \\
    \hdashline
    \multicolumn{11}{c}{\textit{\cellcolor{gray!10}Close-source models}} \\
    ERNIE-4.5-Turbo-VL & 0.189 & 57.25 & 0.356 & 61.38 & 0.429 & 70.98 & 0.369 & 63.95 & 0.334 & 63.97 \\
    Mistral-Large-3 & 0.123 & 51.51 & 0.283 & 61.61 & 0.363 & 67.85 & 0.508 & 75.07 & 0.326 & 64.84 \\
    Claude-3.7-Sonnet & 0.397 & 67.41 & 0.430 & 68.49 & 0.555 & 75.56 & \underline{0.609} & \underline{79.10} & 0.478 & 72.97 \\
    Doubao-Seed-1.6-Vision & 0.332 & 62.94 & 0.295 & 56.43 & 0.543 & 77.10 & 0.444 & 62.25 & 0.386 & 66.07 \\
    Gemini-2.5-Flash & 0.433 & 71.61 & \underline{0.537} & \underline{76.81} & \underline{0.597} & \underline{79.86} & \textbf{0.645} & \textbf{81.41} & \underline{0.552} & \underline{77.54} \\
    o3 & \textbf{0.483} & \textbf{73.82} & 0.442 & 71.91 & 0.558 & 77.84 & 0.523 & 72.01 & 0.487 & 74.23 \\
    GPT-5 & \underline{0.479} & \underline{73.39} & \textbf{0.558} & \textbf{77.75} & \textbf{0.678} & \textbf{82.58} & 0.608 & 78.52 & \textbf{0.569} & \textbf{78.39} \\
    \bottomrule
  \end{tabular}
  }
\end{table*}

\begin{table*}[hbtp]
  \centering
  \caption{FNR and FPR for all evaluated models under the \textit{WithRef} setting. \textbf{Bold} and \underline{underline} indicate the best and second-best scores among models.}
  \label{tab:fnr_fpr_withref}
  \adjustbox{max width=\textwidth}{
  \begin{tabular}{lcccccccccc}
    \toprule
    \multirow{2}{*}{\textbf{Model}} & \multicolumn{2}{c}{\textbf{Physics}} & \multicolumn{2}{c}{\textbf{Geometry}} & \multicolumn{2}{c}{\textbf{Chart}} & \multicolumn{2}{c}{\textbf{Flowchart}} & \multicolumn{2}{c}{\textbf{overall}}\\
    \cmidrule(lr){2-3}\cmidrule(lr){4-5}\cmidrule(lr){6-7}\cmidrule(lr){8-9}\cmidrule(lr){10-11}
    & FNR $\downarrow$ & FPR $\downarrow$
    & FNR $\downarrow$ & FPR $\downarrow$
    & FNR $\downarrow$ & FPR $\downarrow$
    & FNR $\downarrow$ & FPR $\downarrow$
    & FNR $\downarrow$ & FPR $\downarrow$\\
    \midrule
    \multicolumn{11}{c}{\textit{\cellcolor{gray!10}Open-source models}} \\
    Llama-4-Scout-17B-16E-Instruct & 0.278 & 0.440 & 0.375 & 0.283 & 0.151 & 0.404 & 0.319 & 0.212 & 0.255 & 0.354 \\
    Gemma-3-4B-it & 0.167 & 0.688 & \textbf{0.058} & 0.851 & \textbf{0.000} & 0.795 & \textbf{0.052} & 0.915 & \textbf{0.066} & 0.811 \\
    Gemma-3-27B-it & 0.167 & 0.632 & 0.258 & 0.426 & \underline{0.024} & 0.640 & 0.103 & 0.525 & 0.136 & 0.558 \\
    Qwen2.5-VL-3B-Instruct & 1.000 & \textbf{0.000} & 1.000 & \textbf{0.000} & 0.992 & \textbf{0.000} & 1.000 & \textbf{0.000} & 0.998 & \textbf{0.000} \\
    Qwen2.5-VL-7B-Instruct & 0.778 & \underline{0.056} & 0.983 & \underline{0.007} & 0.310 & 0.391 & 0.276 & 0.348 & 0.581 & 0.206 \\
    Qwen2.5-VL-72B-Instruct & 0.833 & 0.064 & 0.792 & 0.085 & 0.087 & 0.379 & 0.379 & 0.170 & 0.511 & 0.185 \\
    Qianfan-VL-8B & \textbf{0.093} & 0.928 & 0.233 & 0.681 & 0.064 & 0.485 & \underline{0.086} & 0.593 & \underline{0.119} & 0.661 \\
    Qianfan-VL-70B & 0.667 & 0.210 & 0.400 & 0.390 & 0.071 & 0.522 & 0.121 & 0.407 & 0.304 & 0.392 \\
    GLM-4.6V & 0.556 & 0.248 & 0.758 & 0.057 & 0.278 & 0.211 & 0.397 & 0.068 & 0.494 & 0.149 \\
    \hdashline
    \multicolumn{11}{c}{\textit{\cellcolor{gray!10}Close-source models}} \\
    ERNIE-4.5-Turbo-VL & 0.630 & 0.200 & 0.667 & 0.057 & 0.405 & \underline{0.180} & 0.586 & 0.093 & 0.568 & 0.134 \\
    Mistral-Large-3 & 0.750 & 0.152 & 0.592 & 0.156 & 0.429 & 0.217 & 0.319 & 0.178 & 0.517 & 0.178 \\
    Claude-3.7-Sonnet & \underline{0.148} & 0.472 & \underline{0.117} & 0.475 & 0.087 & 0.367 & 0.336 & 0.076 & 0.170 & 0.356 \\
    Doubao-Seed-1.6-Vision & 0.602 & 0.112 & 0.742 & 0.050 & 0.238 & 0.217 & 0.672 & \textbf{0.000} & 0.557 & \underline{0.103} \\
    Gemini-2.5-Flash & 0.278 & 0.288 & 0.271 & 0.194 & 0.222 & \underline{0.180} & 0.285 & 0.085 & 0.263 & 0.188 \\
    o3 & 0.213 & 0.304 & 0.358 & 0.206 & 0.222 & 0.217 & 0.491 & \underline{0.042} & 0.321 & 0.196 \\
    GPT-5 & 0.194 & 0.328 & 0.192 & 0.248 & 0.056 & 0.267 & 0.362 & 0.059 & 0.198 & 0.231 \\
    \bottomrule
  \end{tabular}
  }
\end{table*}

\begin{table*}[hbtp]
  \centering
  \caption{Macro$F1_{err}$ ($\times100$) and Micro$F1_{err}$ ($\times100$) for all evaluated models under the \textit{WithRef} setting. \textbf{Bold} and \underline{underline} indicate the best and second-best scores among models.}
  \label{tab:maf1_mif1_withref}
  \adjustbox{max width=\textwidth}{
  \begin{tabular}{lcccccccccc}
    \toprule
    \multirow{2}{*}{\textbf{Model}} & \multicolumn{2}{c}{\textbf{Physics}} & \multicolumn{2}{c}{\textbf{Geometry}} & \multicolumn{2}{c}{\textbf{Chart}} & \multicolumn{2}{c}{\textbf{Flowchart}} & \multicolumn{2}{c}{\textbf{overall}}\\
    \cmidrule(lr){2-3}\cmidrule(lr){4-5}\cmidrule(lr){6-7}\cmidrule(lr){8-9}\cmidrule(lr){10-11}
    & Macro$F1_{err}$ $\uparrow$ & Micro$F1_{err}$ $\uparrow$
    & Macro$F1_{err}$ $\uparrow$ & Micro$F1_{err}$ $\uparrow$
    & Macro$F1_{err}$ $\uparrow$ & Micro$F1_{err}$ $\uparrow$
    & Macro$F1_{err}$ $\uparrow$ & Micro$F1_{err}$ $\uparrow$
    & Macro$F1_{err}$ $\uparrow$ & Micro$F1_{err}$ $\uparrow$ \\
    \midrule
    \multicolumn{11}{c}{\textit{\cellcolor{gray!10}Open-source models}} \\
    Llama-4-Scout-17B-16E-Instruct & 18.54 & 20.20 & 25.63 & 32.56 & 53.36 & 57.45 & 33.53 & 40.62 & 31.56 & 39.40 \\
    Gemma-3-4B-it & 26.06 & 24.47 & 21.33 & 22.22 & 27.90 & 43.87 & 75.76 & 38.30 & 35.28 & 32.03 \\
    Gemma-3-27B-it & 29.34 & 37.50 & 34.38 & 37.25 & 53.56 & 70.42 & 42.55 & 52.17 & 39.15 & 47.93 \\
    Qwen2.5-VL-3B-Instruct & 16.44 & 22.13 & 23.90 & 28.51 & 47.19 & 45.09 & 22.05 & 30.80 & 26.21 & 32.69 \\
    Qwen2.5-VL-7B-Instruct & 27.27 & 28.57 & 21.27 & 30.35 & 45.33 & 47.52 & 20.39 & 33.33 & 28.10 & 34.41 \\
    Qwen2.5-VL-72B-Instruct & 21.74 & 27.30 & 17.60 & 25.13 & 58.82 & 62.95 & 38.89 & 43.72 & 33.38 & 38.05 \\
    Qianfan-VL-8B & 23.33 & 41.67 & 13.56 & 21.74 & 33.56 & 33.33 & 29.76 & 36.59 & 25.28 & 32.70 \\
    Qianfan-VL-70B & 30.32 & 39.07 & 14.21 & 19.89 & 46.59 & 58.54 & 31.35 & 35.83 & 30.62 & 38.76 \\
    GLM-4.6V & 35.19 & 46.01 & 34.31 & 37.95 & 61.19 & 66.20 & 50.26 & 56.28 & 44.58 & 51.72 \\
    \hdashline
    \multicolumn{11}{c}{\textit{\cellcolor{gray!10}Close-source models}} \\
    ERNIE-4.5-Turbo-VL & 33.05 & 36.97 & 22.22 & 27.34 & 51.95 & 50.45 & 37.04 & 37.25 & 35.84 & 38.25 \\
    Mistral-Large-3 & 19.89 & 29.00 & 30.61 & 32.98 & 53.76 & 55.14 & 38.02 & 46.24 & 34.31 & 41.04\\
    Claude-3.7-Sonnet & 34.31 & \underline{47.06} & 32.56 & 37.32 & 57.11 & 66.67 & 52.52 & 60.42 & 43.64 & 54.49 \\
    Doubao-Seed-1.6-Vision & \underline{40.52} & 45.49 & 30.46 & 36.36 & \underline{64.28} & \underline{69.13} & 56.86 & 60.54 & 47.76 & 53.16 \\
    Gemini-2.5-Flash & \textbf{41.30} & \textbf{51.14} & 42.67 & \textbf{44.83} & 60.23 & 63.26 & 55.84 & 61.44 & \underline{49.50} & \textbf{55.74} \\
    o3 & 33.33 & 40.57 & \underline{42.97} & \underline{42.90} & 59.07 & 63.87 & \underline{57.44} & \underline{61.54} & 47.31 & 53.23 \\
    GPT-5 & 34.79 & 43.88 & \textbf{44.06} & 43.70 & \textbf{66.05} & \textbf{73.19} & \textbf{58.52} & \textbf{62.25} & \textbf{49.79} & \underline{55.71} \\
    \bottomrule
  \end{tabular}
  }
\end{table*}

\begin{table*}[hbtp]
  \centering
  \caption{Acc and eb$F_1$ for all evaluated models under the \textit{NoRef} setting. \textbf{Bold} and \underline{underline} indicate the best and second-best scores among models.}
  \label{tab:noref_results}
  \adjustbox{max width=\textwidth}{
  \begin{tabular}{lcccccccccc}
    \toprule
    \multirow{2}{*}{\textbf{Model}} & \multicolumn{2}{c}{\textbf{Physics}} & \multicolumn{2}{c}{\textbf{Geometry}} & \multicolumn{2}{c}{\textbf{Chart}} & \multicolumn{2}{c}{\textbf{Flowchart}} & \multicolumn{2}{c}{\textbf{overall}}\\
    \cmidrule(lr){2-3}\cmidrule(lr){4-5}\cmidrule(lr){6-7}\cmidrule(lr){8-9}\cmidrule(lr){10-11}
    & Acc $\uparrow$ & eb$F_1 \uparrow$
    & Acc $\uparrow$ & eb$F_1 \uparrow$
    & Acc $\uparrow$ & eb$F_1 \uparrow$
    & Acc $\uparrow$ & eb$F_1 \uparrow$
    & Acc $\uparrow$ & eb$F_1 \uparrow$\\
    \midrule
    Random choice & 50.00 & -- & 50.00 & -- & 50.00 & -- & 50.00 & -- & 50.00 & -- \\
    \hline
    \multicolumn{11}{c}{\textit{\cellcolor{gray!10}Open-source models}} \\
    Llama-4-Scout-17B-16E-Instruct & 56.65 & 21.59 & 57.09 & 23.29 & 68.64 & 60.10 & 59.83 & 37.88 & 60.89 & 38.31 \\
    Gemma-3-4B-it & 52.36 & 29.28 & 46.74 & 26.97 & 49.13 & 43.10 & 58.12 & 31.46 & 51.33 & 31.95 \\
    Gemma-3-27B-it & 55.79 & 31.28 & 55.94 & 28.44 & 60.63 & 59.09 & \underline{70.51} & 44.21 & 60.59 & 39.44\\
    Qwen2.5-VL-3B-Instruct & 53.65 & 18.40 & 54.02 & 20.85 & 56.10 & 38.59 & 50.43 & 30.59 & 53.69 & 27.64 \\
    Qwen2.5-VL-7B-Instruct & 57.08 & 24.85 & 54.41 & 26.07 & 61.32 & 39.44 & 66.24 & 34.86 & 59.70 & 31.13 \\
    Qwen2.5-VL-72B-Instruct & 56.65 & 27.78 & 54.41 & 26.51 & 68.29 & 58.61 & 64.96 & 41.84 & 61.28 & 38.65 \\
    Qianfan-VL-8B & 50.43 & 35.56 & 51.34 & 24.56 & 55.75 & 32.86 & 58.97 & 47.54 & 54.14 & 35.76 \\
    Qianfan-VL-70B & 53.88 & 36.43 & 45.59 & 18.75 & 61.32 & 50.81 & 62.82 & 36.94 & 55.92 & 35.70 \\
    GLM-4.6V & 55.79 & \textbf{40.76} & 59.00 & 33.17 & 72.82 & 58.94 & \textbf{74.36} & \textbf{60.07} & 65.71 & 48.61 \\
    \hdashline
    \multicolumn{11}{c}{\textit{\cellcolor{gray!10}Closed-source models}} \\
    ERNIE-4.5-Turbo-VL & 54.94 & 35.36 & 52.49 & 27.82 & 59.93 & 44.43 & 53.42 & 35.33 & 55.37 & 35.55 \\
    Mistral-Large-3 & 56.22 & 34.73 & 52.11 & 28.10 & 67.94 & 51.90 & 61.54 & 43.56 & 59.70 & 40.43 \\
    Claude-3.7-Sonnet & 59.66 & \underline{39.79} & 56.32 & 31.42 & 64.46 & 57.72 & 56.41 & 52.63 & 59.41 & 46.50 \\
    Doubao-Seed-1.6-Vision & 57.08 & 36.42 & 61.30 & \underline{36.53} & 73.87 & 59.93 & 61.97 & 56.86 & 64.04 & 47.90 \\
    Gemini-2.5-Flash & \underline{60.52} & 37.41 & 63.22 & 35.19 & 73.52 & 62.56 & 69.23 & 57.50 & 66.90 & \underline{49.46} \\
    o3 & 59.23 & 35.26 & \textbf{68.97} & 34.87 & \textbf{78.75} & \underline{65.16} & 63.25 & \underline{58.83} & \underline{68.18} & 49.15 \\
    GPT-5 & \textbf{64.38} & 35.88 & \underline{68.20} & \textbf{41.07} & \underline{78.40} & \textbf{71.84} & 67.52 & 58.72 & \textbf{70.05} & \textbf{52.38}\\
    \bottomrule
  \end{tabular}
  }
\end{table*}

\subsection{Grading Without Reference Answers}

Table~\ref{tab:noref_results} reports \textbf{Acc} and \textbf{eb$F_1$} for all evaluated models under the \textit{NoRef} setting, where reference diagrams are removed and models must judge correctness solely based on the problem statement and the student sketch. Overall performance drops substantially compared to the \textit{WithRef} setting, suggesting that reference-free grading requires \textbf{solving the underlying task} in addition to \textbf{interpreting noisy hand-drawn diagrams}. Although the random baseline yields 50\% accuracy, most models only reach the low-to-mid 60\% range, indicating that reliable grading remains challenging when models must jointly perform problem solving, diagram understanding, and robustness to freehand variation. Consistent with the main results, closed-source models remain stronger, but the gap between models narrows under \textit{NoRef}, highlighting the increased difficulty of this setting.

\subsection{Hand-Drawn vs.\ Electronic Sketches}
\label{app:hand_vs_electronic}

\begin{table*}[hbtp]
  \centering
  \small
  \caption{\textbf{Performance Comparison on Hand-Drawn vs.\ Electronic Sketches.} Accuracy (Acc) and example-based F1 score (eb$F_1$) for hand-drawn and electronic sketches. The \textit{Delta} column shows the difference (Electronic $-$ Hand-drawn). All model results are reported in the \textit{WithRef} setting. \textbf{Bold} and \underline{underline} indicate the best and second-best scores within the \textit{Hand-Drawn} and \textit{Electronic} columns, respectively. For the \textit{Delta} column, positive values are shown in \textcolor{green!60!black}{green} and negative values in \textcolor{red!70!black}{red}.}
  \label{tab:hand_vs_electronic}
  \setlength{\tabcolsep}{9pt}
  \begin{tabular}{L{4.3cm}C{1.2cm}C{1.2cm}C{1.2cm}C{1.2cm}C{1.2cm}C{1.2cm}C{1.2cm}C{1.2cm}}
    \toprule
    \multirow{2}{*}{\textbf{Model}} & \multicolumn{2}{c}{\textbf{Hand-Drawn}} & \multicolumn{2}{c}{\textbf{Electronic}} & \multicolumn{2}{c}{\textbf{Difference}} \\
    \cmidrule(lr){2-7}
    & \textbf{Acc} & \textbf{eb$F_1$} & \textbf{Acc} & \textbf{eb$F_1$} & \textbf{$\Delta$ Acc} & \textbf{$\Delta$ eb$F_1$} \\
    \midrule
    \multicolumn{7}{c}{\textit{\cellcolor{gray!10}Open-source models}} \\
    Llama-4-Scout-17B-16E-Instruct & 69.78 & 39.91 & 67.21 & 38.80 & \textcolor{red!70!black}{-2.57} & \textcolor{red!70!black}{-1.12} \\
    Gemma-3-4B-it & 53.83 & 32.20 & 52.05 & 29.55 & \textcolor{red!70!black}{-1.78} & \textcolor{red!70!black}{-2.65} \\
    Gemma-3-27B-it & 64.98 & 49.20 & 59.84 & 49.62 & \textcolor{red!70!black}{-5.14} & \textcolor{green!60!black}{+0.42} \\
    Qwen2.5-VL-3B-Instruct & 53.96 & 32.21 & 53.28 & 29.72 & \textcolor{red!70!black}{-0.68} & \textcolor{red!70!black}{-2.50} \\
    Qwen2.5-VL-7B-Instruct & 60.70 & 35.80 & 66.39 & 28.00 & \textcolor{green!60!black}{+5.69} & \textcolor{red!70!black}{-7.80} \\
    Qwen2.5-VL-72B-Instruct & 66.41 & 40.34 & 66.39 & 35.72 & \textcolor{red!70!black}{-0.01} & \textcolor{red!70!black}{-4.62} \\
    Qianfan-VL-8B & 59.53 & 31.47 & 57.38 & 36.38 & \textcolor{red!70!black}{-2.16} & \textcolor{green!60!black}{+4.91} \\
    Qianfan-VL-70B & 65.50 & 39.63 & 62.96 & 37.65 & \textcolor{red!70!black}{-2.54} & \textcolor{red!70!black}{-1.97} \\
    GLM-4.6V & 69.26 & 52.73 & 68.85 & 55.44 & \textcolor{red!70!black}{-0.41} & \textcolor{green!60!black}{+2.71} \\
    \hdashline
    \multicolumn{7}{c}{\textit{\cellcolor{gray!10}Closed-source models}} \\
    ERNIE-4.5-Turbo-VL & 65.89 & 38.37 & 68.44 & 39.19 & \textcolor{green!60!black}{+2.55} & \textcolor{green!60!black}{+0.82} \\
    Mistral-Large-3 & 67.19 & 42.93 & 64.34 & 38.89 & \textcolor{red!70!black}{-2.84} & \textcolor{red!70!black}{-4.04} \\
    Claude-3.7-Sonnet & 73.28 & \underline{56.86} & 72.13 & 57.96 & \textcolor{red!70!black}{-1.15} & \textcolor{green!60!black}{+1.09} \\
    Doubao-Seed-1.6-Vision & 68.48 & 54.63 & 69.26 & 54.11 & \textcolor{green!60!black}{+0.78} & \textcolor{red!70!black}{-0.52} \\
    Gemini-2.5-Flash & \underline{77.18} & 56.85 & \textbf{79.51} & \textbf{62.94} & \textcolor{green!60!black}{+2.32} & \textcolor{green!60!black}{+6.09} \\
    o3 & 74.19 & 55.84 & 75.82 & 56.17 & \textcolor{green!60!black}{+1.63} & \textcolor{green!60!black}{+0.33} \\
    GPT-5 & \textbf{78.21} & \textbf{60.97} & \underline{79.10} & \underline{61.65} & \textcolor{green!60!black}{+0.89} & \textcolor{green!60!black}{+0.68} \\
    \bottomrule
  \end{tabular}
\end{table*}

\begin{table*}[htbp]
  \centering
  \small
  \caption{\textbf{Exact-match vs.\ alternative correct solutions.} Accuracy (Acc) on two subsets of \textbf{correct} sketches: (i) sketches closely matching the reference diagram (\textit{Exact}), and (ii) sketches that remain correct but deviate from the reference in structure or strategy (\textit{Alt}). We also report the gap $\Delta = \textit{Exact} - \textit{Alt}$. All model accuracies are reported in the \textit{WithRef} setting. \textbf{Bold} and \underline{underline} indicate the best and second-best scores in each column. For $\Delta$, positive gaps are shown in \textcolor{green!60!black}{green} and negative gaps in \textcolor{red!70!black}{red}.}
  \label{tab:exact_alt_acc}
  \setlength{\tabcolsep}{4.5pt}
  \renewcommand{\arraystretch}{1.12}
  \adjustbox{max width=\textwidth}{
  \begin{tabular}{lccc ccc ccc ccc ccc}
    \toprule
    \multirow{2}{*}{\textbf{Model}} & \multicolumn{3}{c}{\textbf{Physics}} & \multicolumn{3}{c}{\textbf{Geometry}} & \multicolumn{3}{c}{\textbf{Chart}} & \multicolumn{3}{c}{\textbf{Flowchart}} & \multicolumn{3}{c}{\textbf{Overall}} \\
    \cmidrule(lr){2-4}\cmidrule(lr){5-7}\cmidrule(lr){8-10}\cmidrule(lr){11-13}\cmidrule(lr){14-16}
    & Exact & Alt & $\Delta$ & Exact & Alt & $\Delta$ & Exact & Alt & $\Delta$ & Exact & Alt & $\Delta$ & Exact & Alt & $\Delta$ \\
    \midrule
    \multicolumn{16}{c}{\textit{\cellcolor{gray!10}Open-source models}} \\
    Llama-4-Scout-17B-16E-Instruct & 74.19 & 60.00 & \textcolor{green!60!black}{+14.19} & 76.92 & 55.17 & \textcolor{green!60!black}{+21.75} & 85.42 & 83.33 & \textcolor{green!60!black}{+2.08} & 71.13 & 52.63 & \textcolor{green!60!black}{+18.50} & 76.92 & 64.52 & \textcolor{green!60!black}{+12.41} \\
    Gemma-3-4B-it & 83.87 & \underline{80.00} & \textcolor{green!60!black}{+3.87} & \textbf{92.31} & \textbf{100.00} & \textcolor{red!70!black}{-7.69} & \textbf{100.00} & \textbf{100.00} & 0.00 & \textbf{95.88} & \textbf{89.47} & \textcolor{green!60!black}{+6.40} & \textbf{93.10} & \textbf{94.62} & \textcolor{red!70!black}{-1.52} \\
    Gemma-3-27B-it & 83.87 & \underline{80.00} & \textcolor{green!60!black}{+3.87} & 84.62 & 41.38 & \textcolor{green!60!black}{+43.24} & \underline{98.96} & 93.33 & \textcolor{green!60!black}{+5.62} & \underline{93.81} & 68.42 & \textcolor{green!60!black}{+25.39} & 90.45 & 69.89 & \textcolor{green!60!black}{+20.56} \\
    Qwen2.5-VL-3B-Instruct & 0.00 & 0.00 & 0.00 & 0.00 & 0.00 & 0.00 & 1.04 & 0.00 & \textcolor{green!60!black}{+1.04} & 0.00 & 0.00 & 0.00 & 0.27 & 0.00 & \textcolor{green!60!black}{+0.27} \\
    Qwen2.5-VL-7B-Instruct & 23.66 & 13.33 & \textcolor{green!60!black}{+10.32} & 2.20 & 0.00 & \textcolor{green!60!black}{+2.20} & 69.79 & 66.67 & \textcolor{green!60!black}{+3.12} & 77.32 & 47.37 & \textcolor{green!60!black}{+29.95} & 44.03 & 33.33 & \textcolor{green!60!black}{+10.70} \\
    Qwen2.5-VL-72B-Instruct & 16.13 & 20.00 & \textcolor{red!70!black}{-3.87} & 23.08 & 13.79 & \textcolor{green!60!black}{+9.28} & 92.71 & 86.67 & \textcolor{green!60!black}{+6.04} & 67.01 & 36.84 & \textcolor{green!60!black}{+30.17} & 50.40 & 43.01 & \textcolor{green!60!black}{+7.39} \\
    Qianfan-VL-8B & \underline{91.40} & \textbf{86.67} & \textcolor{green!60!black}{+4.73} & 83.52 & 55.17 & \textcolor{green!60!black}{+28.34} & 94.79 & 90.00 & \textcolor{green!60!black}{+4.79} & \underline{93.81} & \underline{78.95} & \textcolor{green!60!black}{+14.87} & \underline{90.98} & \underline{76.34} & \textcolor{green!60!black}{+14.64} \\
    Qianfan-VL-70B & 32.26 & 40.00 & \textcolor{red!70!black}{-7.74} & 64.84 & 44.83 & \textcolor{green!60!black}{+20.01} & 93.75 & 90.00 & \textcolor{green!60!black}{+3.75} & 92.78 & 63.16 & \textcolor{green!60!black}{+29.63} & 71.35 & 62.37 & \textcolor{green!60!black}{+8.99} \\
    GLM-4.6V & 45.16 & 40.00 & \textcolor{green!60!black}{+5.16} & 27.47 & 13.79 & \textcolor{green!60!black}{+13.68} & 71.88 & 73.33 & \textcolor{red!70!black}{-1.45} & 64.95 & 36.84 & \textcolor{green!60!black}{+28.11} & 52.79 & 41.94 & \textcolor{green!60!black}{+10.85} \\
    \hdashline
    \multicolumn{16}{c}{\textit{\cellcolor{gray!10}Closed-source models}} \\
    ERNIE-4.5-Turbo-VL & 37.63 & 33.33 & \textcolor{green!60!black}{+4.30} & 40.66 & 10.34 & \textcolor{green!60!black}{+30.31} & 55.21 & 73.33 & \textcolor{red!70!black}{-18.12} & 47.42 & 10.53 & \textcolor{green!60!black}{+36.90} & 45.36 & 34.41 & \textcolor{green!60!black}{+10.95} \\
    Mistral-Large-3 & 23.66 & 33.33 & \textcolor{red!70!black}{-9.68} & 47.25 & 20.69 & \textcolor{green!60!black}{+26.56} & 59.38 & 50.00 & \textcolor{green!60!black}{+9.38} & 74.23 & 36.84 & \textcolor{green!60!black}{+37.39} & 51.46 & 35.48 & \textcolor{green!60!black}{+15.98} \\
    Claude-3.7-Sonnet & \textbf{92.47} & 40.00 & \textcolor{green!60!black}{+52.47} & \textbf{92.31} & \underline{75.86} & \textcolor{green!60!black}{+16.45} & 95.83 & 76.67 & \textcolor{green!60!black}{+19.16} & 74.23 & 26.32 & \textcolor{green!60!black}{+47.91} & 88.59 & 60.22 & \textcolor{green!60!black}{+28.38} \\
    Doubao-Seed-1.6-Vision & 40.86 & 33.33 & \textcolor{green!60!black}{+7.53} & 26.37 & 24.14 & \textcolor{green!60!black}{+2.24} & 76.04 & 76.67 & \textcolor{red!70!black}{-0.63} & 36.08 & 15.79 & \textcolor{green!60!black}{+20.29} & 45.09 & 40.86 & \textcolor{green!60!black}{+4.23} \\
    Gemini-2.5-Flash & 73.12 & 66.67 & \textcolor{green!60!black}{+6.45} & 78.65 & 55.17 & \textcolor{green!60!black}{+23.48} & 80.21 & 70.00 & \textcolor{green!60!black}{+10.21} & 73.20 & 63.16 & \textcolor{green!60!black}{+10.04} & 76.27 & 63.44 & \textcolor{green!60!black}{+12.83} \\
    o3 & 82.80 & 53.33 & \textcolor{green!60!black}{+29.46} & 71.43 & 41.38 & \textcolor{green!60!black}{+30.05} & 79.17 & 73.33 & \textcolor{green!60!black}{+5.84} & 50.52 & 52.63 & \textcolor{red!70!black}{-2.12} & 70.82 & 55.91 & \textcolor{green!60!black}{+14.91} \\
    GPT-5 & 82.80 & 66.67 & \textcolor{green!60!black}{+16.13} & \underline{87.91} & 58.62 & \textcolor{green!60!black}{+29.29} & 93.75 & \underline{96.67} & \textcolor{red!70!black}{-2.92} & 62.89 & 68.42 & \textcolor{red!70!black}{-5.53} & 81.70 & 74.19 & \textcolor{green!60!black}{+7.51} \\
    \bottomrule
  \end{tabular}}
\end{table*}

In addition to structural variation in correct solutions, we further investigate whether model performance depends on the \textbf{sketch modality}. Specifically, we compare results on \textbf{hand-drawn} sketches (captured from freehand writing) versus \textbf{electronic} sketches (produced using stylus or digital tools). Our benchmark contains \textbf{771 hand-drawn} sketches and \textbf{244 electronic} sketches, reflecting the naturally higher prevalence of freehand student solutions.

Table~\ref{tab:hand_vs_electronic} reports \textbf{Acc} and \textbf{eb$F_1$} on the two modalities. Overall, performance differences are relatively small across most models, suggesting that modern MLLMs are broadly robust to the style and rendering differences between hand-drawn and electronic sketches. While some models achieve slightly higher accuracy on electronic sketches (e.g., Gemini-2.5-Flash and GPT-5), the average gap is modest and not consistently positive across models. In fact, several models exhibit small negative deltas, indicating that the cleaner appearance of electronic sketches does not always translate into easier grading.

\subsection{Exact vs.\ Alternative Correct Answers}
\label{app:exact_vs_alt}

To examine whether models can verify \emph{semantic correctness} beyond strict reference matching, we further divide all \textbf{correct} student answers into two subgroups: (i) \textbf{reference-aligned correct} answers that closely follow the canonical reference diagram, and (ii) \textbf{reference-divergent correct} answers that remain valid but differ from the reference (e.g., through structural deformation, alternative construction strategies, or equivalent circuit layouts). Since both subgroups are correct, we report only correctness-level performance (Acc) in this analysis.

Table~\ref{tab:exact_alt_dist} summarizes the distribution of the two correct-answer subtypes across domains. As expected, most correct answers are reference-aligned, and alternative correct solutions are less frequent. Nevertheless, the \textit{Alt} subset remains sizable for meaningful comparison.

\begin{table}[hbtp]
  \centering
  \small
  \caption{\textbf{Distribution of correct-answer subtypes.} Counts and proportions of reference-aligned (\textit{Exact}) and reference-divergent (\textit{Alt}) correct sketches in each domain.}
  \label{tab:exact_alt_dist}
  \renewcommand{\arraystretch}{1.0}
  \begin{tabular}{L{1.2cm}R{0.9cm}R{0.8cm}C{1.5cm}C{1.3cm}}
    \toprule
    \textbf{Domain} & \textbf{\#Exact} & \textbf{\#Alt} & \textbf{Exact (\%)} & \textbf{Alt (\%)} \\
    \midrule
    Physics   & 93  & 15  & 86.1 & 13.9 \\
    Geometry  & 91  & 29  & 75.8 & 24.2 \\
    Chart     & 96  & 30  & 76.2 & 23.8 \\
    Flowchart & 97  & 19  & 83.6 & 16.4 \\
    \midrule
    Overall   & 377 & 93  & 80.2 & 19.8 \\
    \bottomrule
  \end{tabular}
\end{table}

Table~\ref{tab:exact_alt_acc} compares model accuracy on these two subgroups. Most models achieve consistently higher accuracy on reference-aligned answers, while performance drops noticeably on reference-divergent but still correct solutions. This effect is especially pronounced in Physics and Flowchart, where several strong models exhibit large gaps (e.g., Claude-3.7-Sonnet: \textcolor{green!60!black}{$52.47\uparrow$} in Physics and \textcolor{green!60!black}{$47.91\uparrow$} in Flowchart; o3: \textcolor{green!60!black}{$29.46\uparrow$} in Physics). In contrast, the performance on Chart shows smaller or even negative gaps for some models (e.g., Gemma-3-4B-it and GPT-5), suggesting that chart correctness is less sensitive to stylistic or structural variation.

A plausible explanation is that reference-divergent solutions require models to abstract away from surface-level visual similarity and instead verify \emph{structural equivalence} (e.g., topological consistency in flowcharts or physically equivalent circuit layouts), which remains challenging under freehand noise. By comparison, charts follow more rigid graphical conventions, so alternative correct forms often preserve key geometric cues (axis alignment, bar/line placement), reducing the reliance on canonical matching.

\subsection{Case Studies of Model Responses}
\label{app:case_study}

We provide qualitative examples that visualize \textbf{model grading responses} alongside the gold annotations, highlighting where models succeed, where they fail, and how different error types manifest in real student sketches

\begin{figure*}[htbp]
    \centering
    \includegraphics[width=\linewidth]{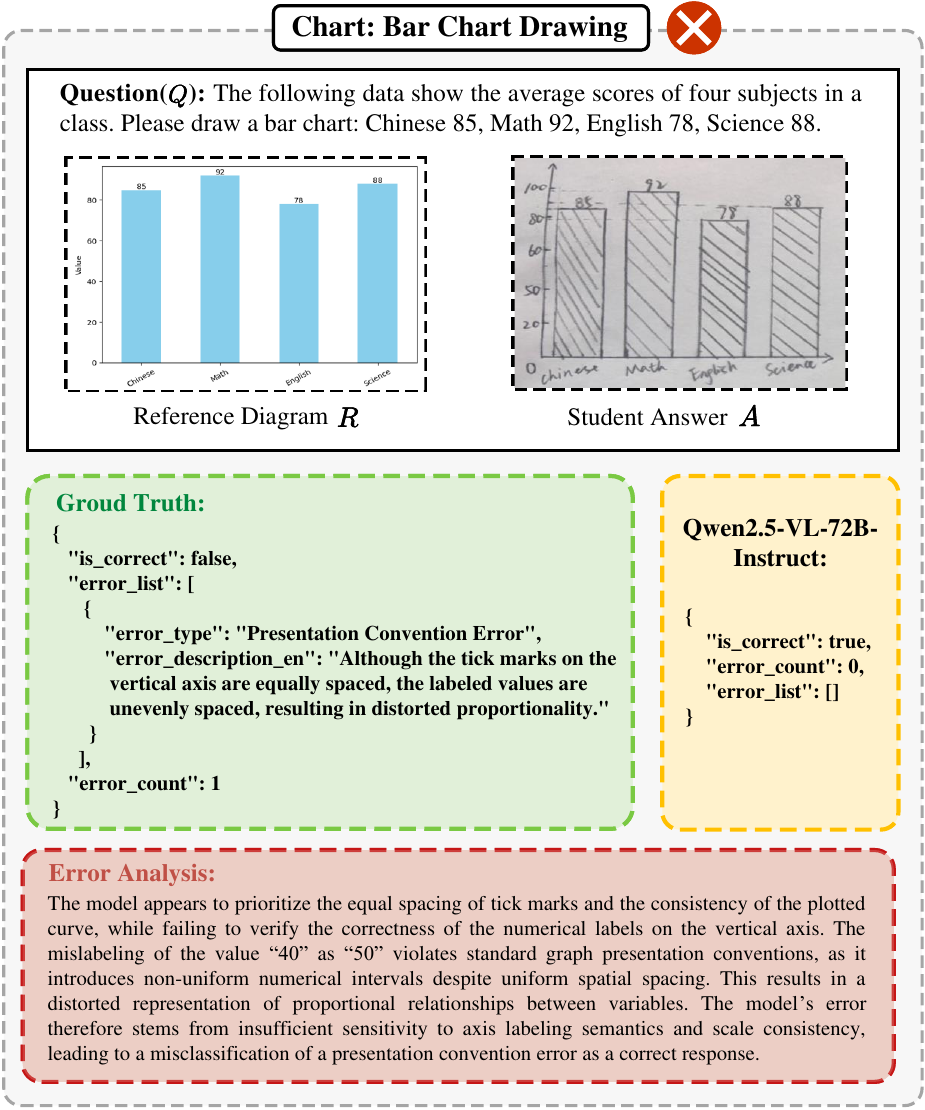}
    \caption{A sample error case in chart.}
\end{figure*}

\begin{figure*}[htbp]
    \centering
    \includegraphics[width=\linewidth]{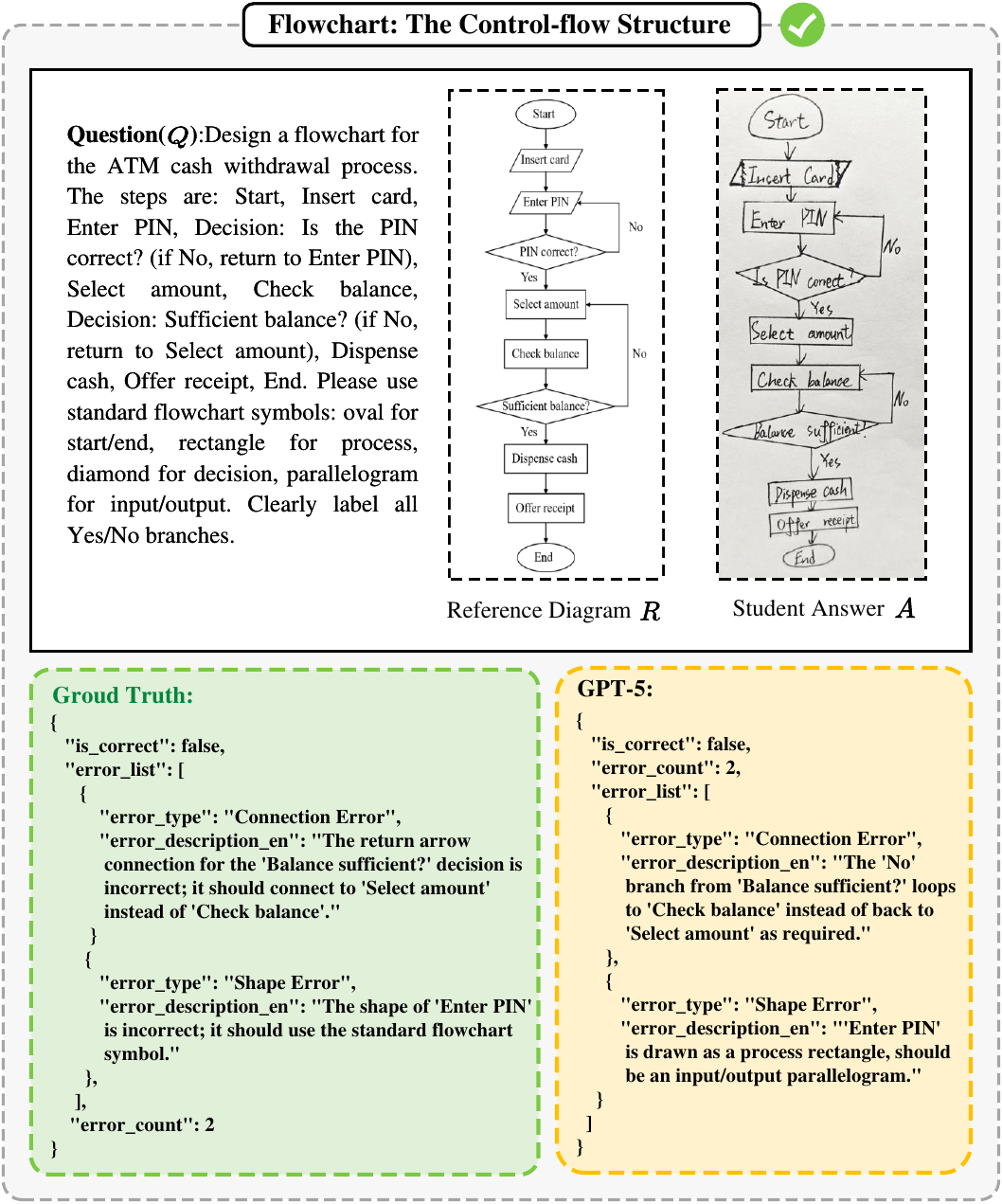}
    \caption{A sample correct case in flowchart.}
\end{figure*}

\begin{figure*}[htbp]
    \centering
    \includegraphics[width=\linewidth]{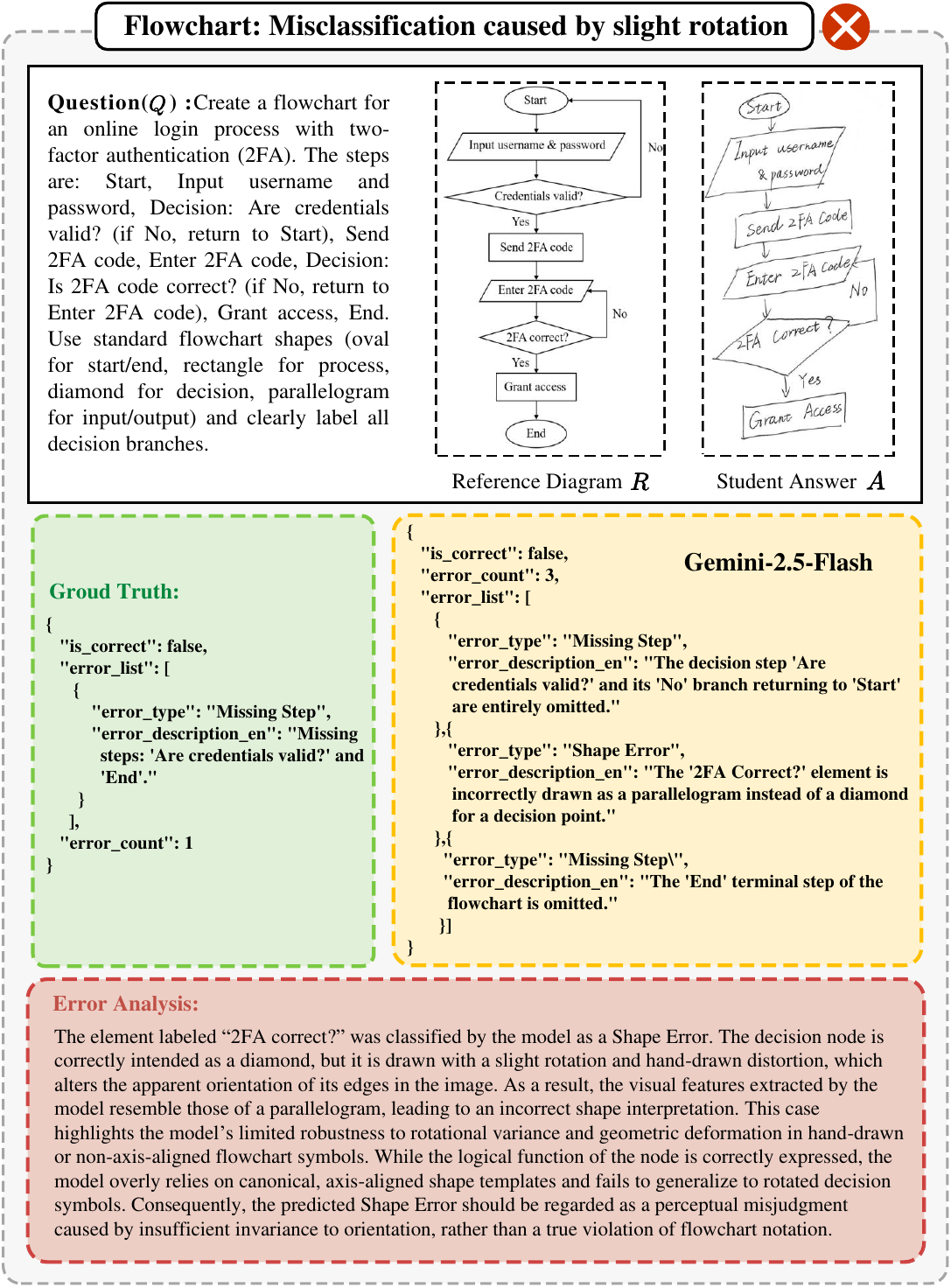}
    \caption{A sample error case in flowchart.}
\end{figure*}

\begin{figure*}[htbp]
    \centering
    \includegraphics[width=\linewidth]{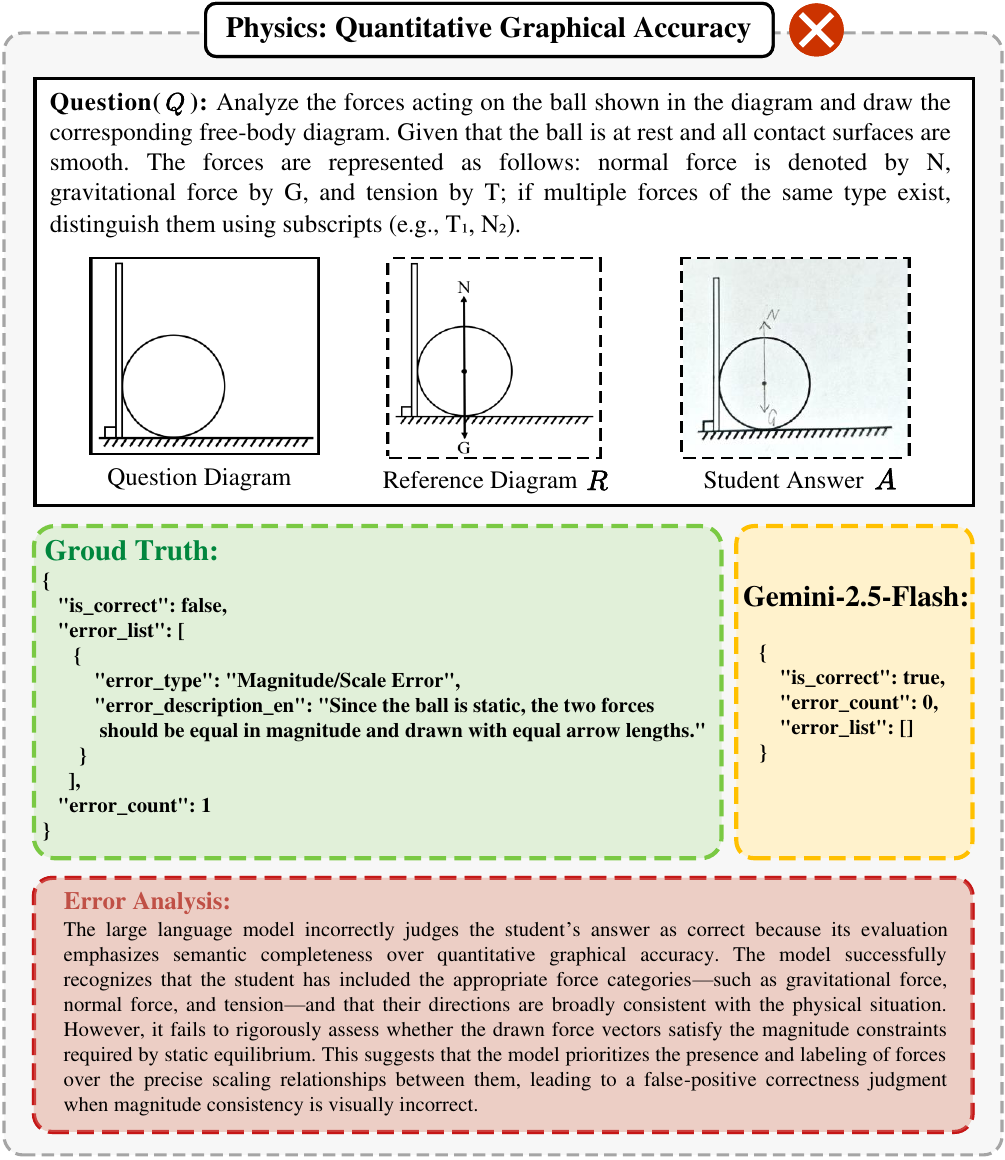}
    \caption{A sample error case in physics.}
\end{figure*}

\begin{figure*}[htbp]
    \centering
    \includegraphics[width=0.92\linewidth]{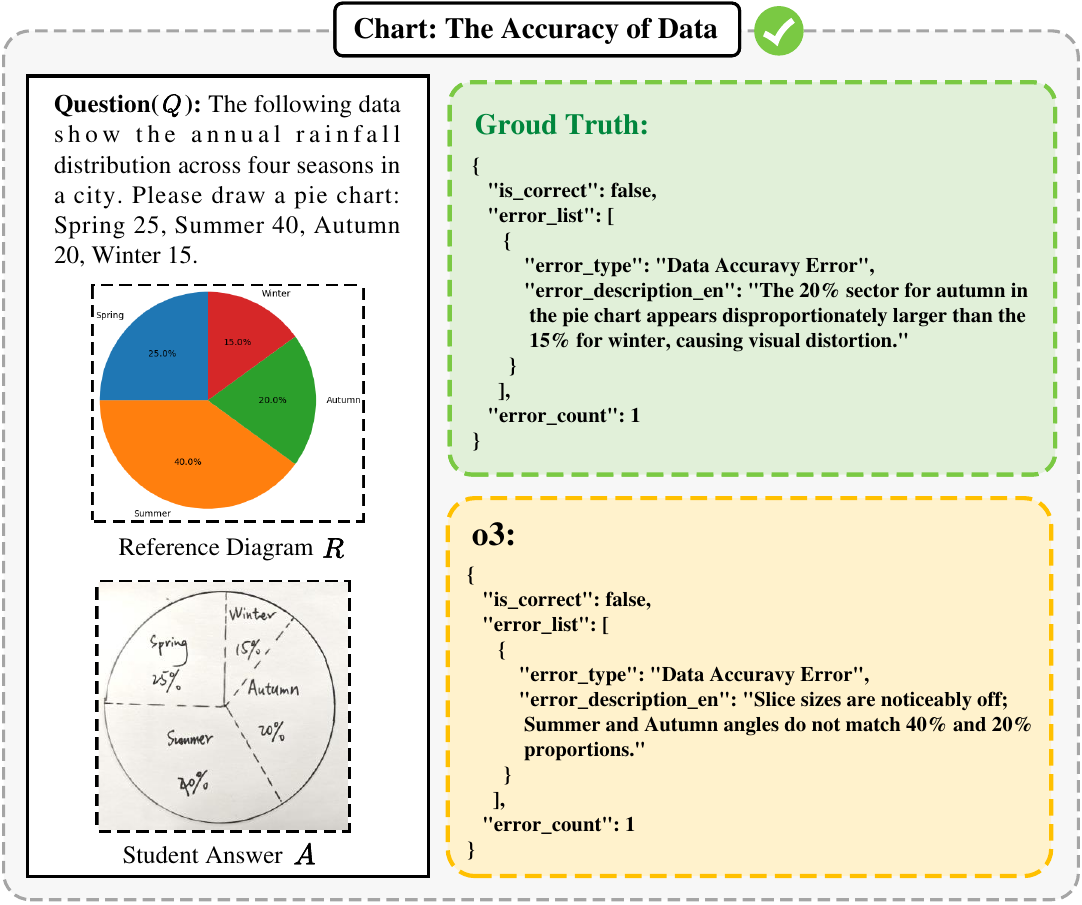}
    \caption{A sample correct case in chart.}
\end{figure*}

\begin{figure*}[htbp]
    \centering
    \includegraphics[width=0.92\linewidth]{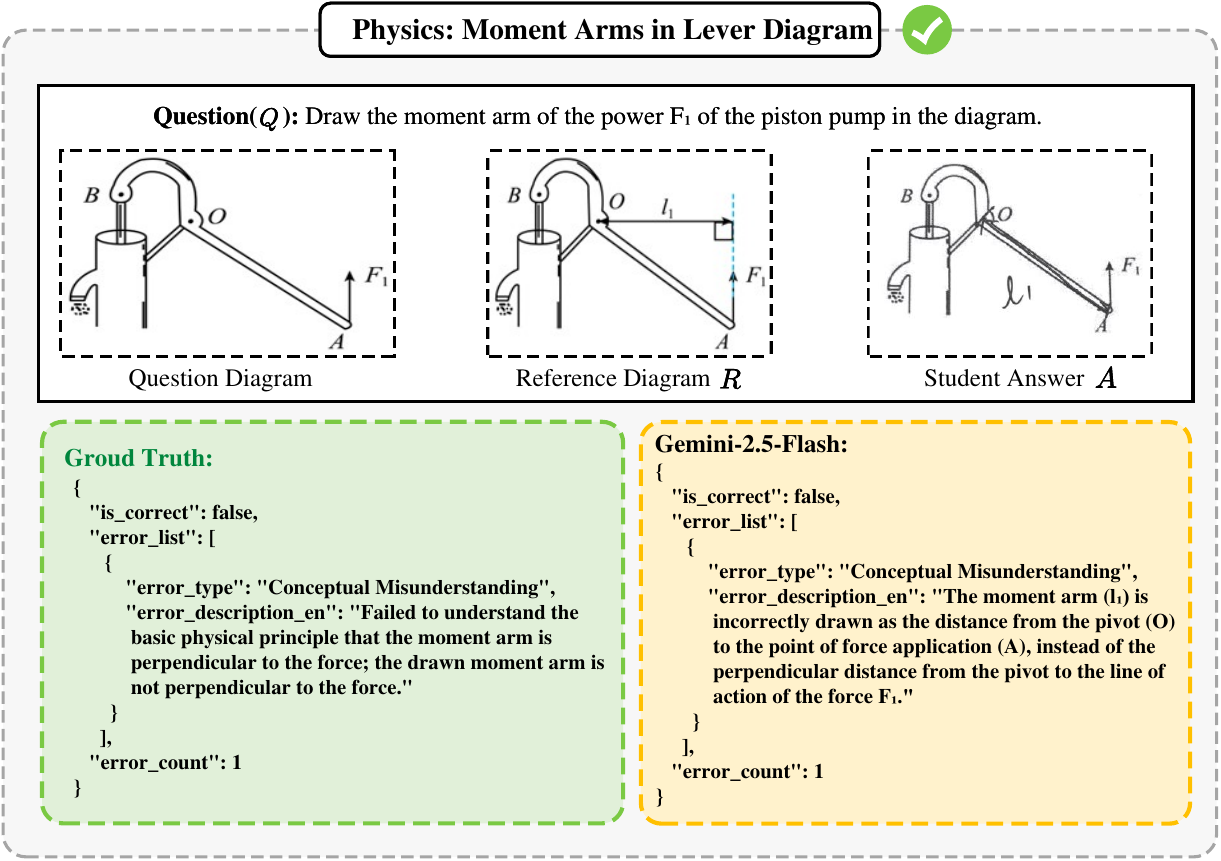}
    \caption{A sample correct case in physics.}
\end{figure*}

\begin{figure*}[htbp]
    \centering
    \includegraphics[width=\linewidth]{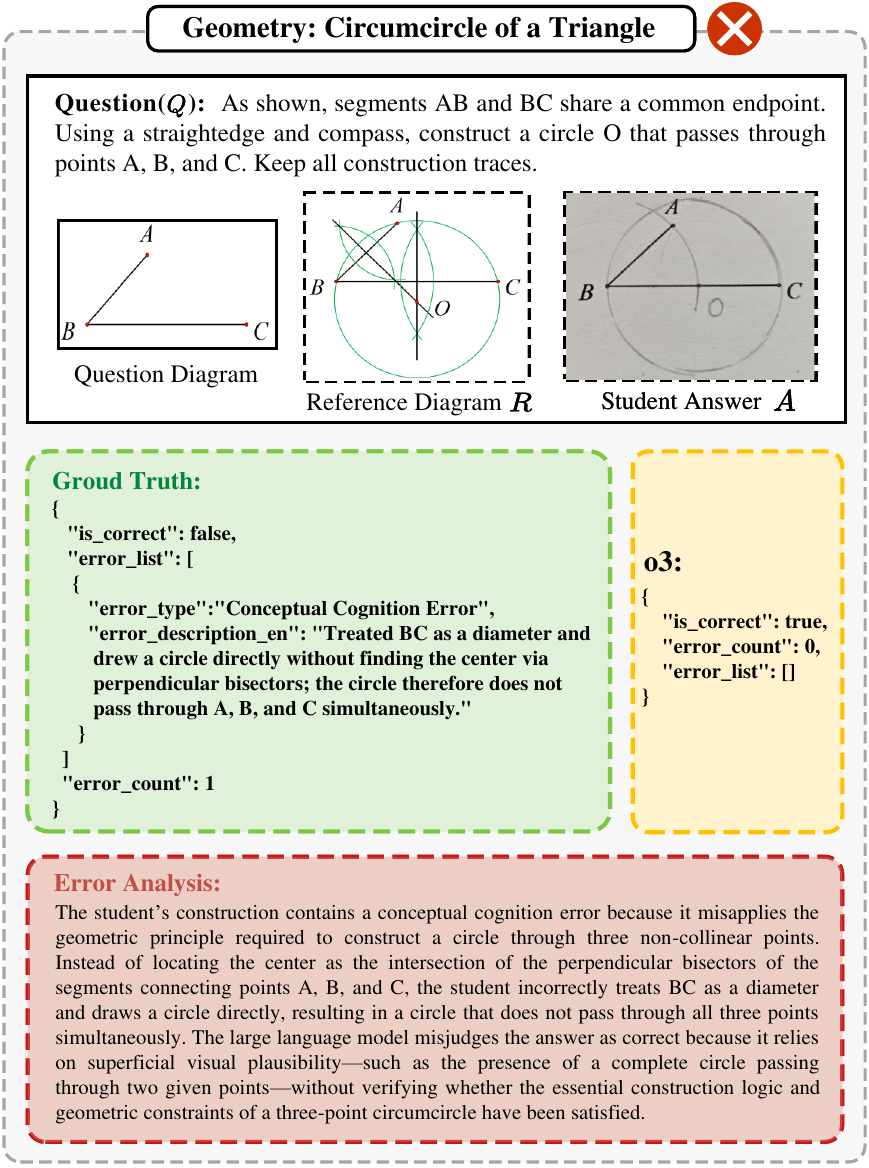}
    \caption{A sample error case in geometry.}
\end{figure*}

\begin{figure*}[htbp]
    \centering
    \includegraphics[width=\linewidth]{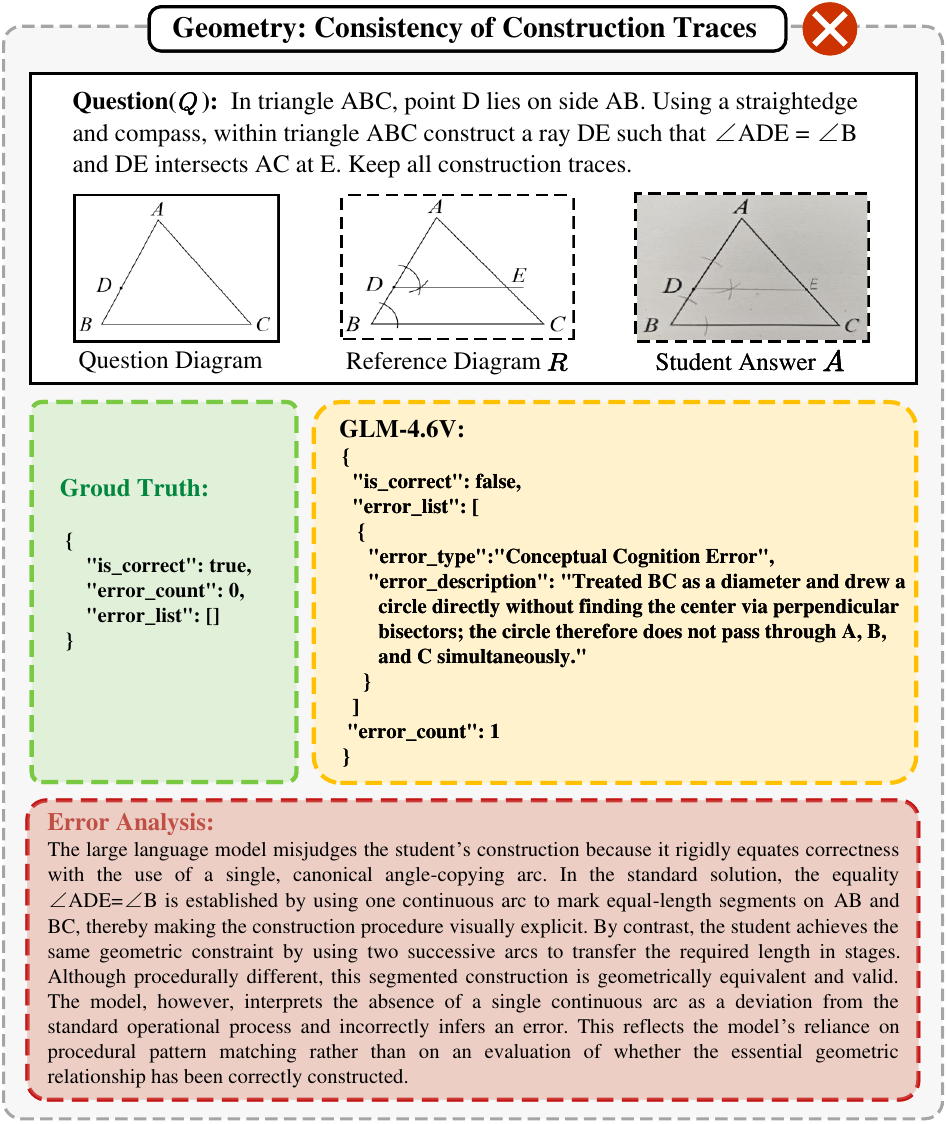}
    \caption{A sample error case in geometry.}
\end{figure*}

\begin{figure*}[htbp]
    \centering
    \includegraphics[width=\linewidth]{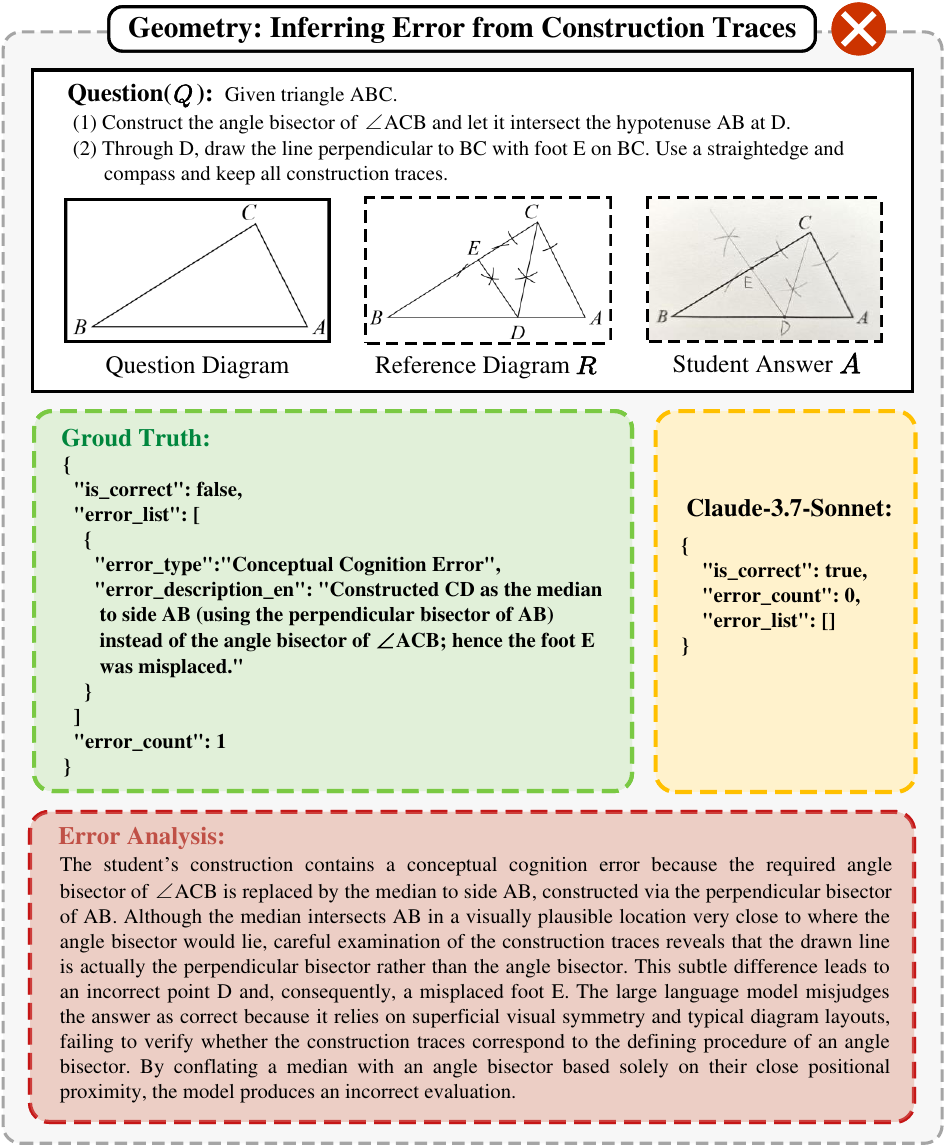}
    \caption{A sample error case in geometry.}
\end{figure*}

\begin{figure*}[htbp]
    \centering
    \includegraphics[width=\linewidth]{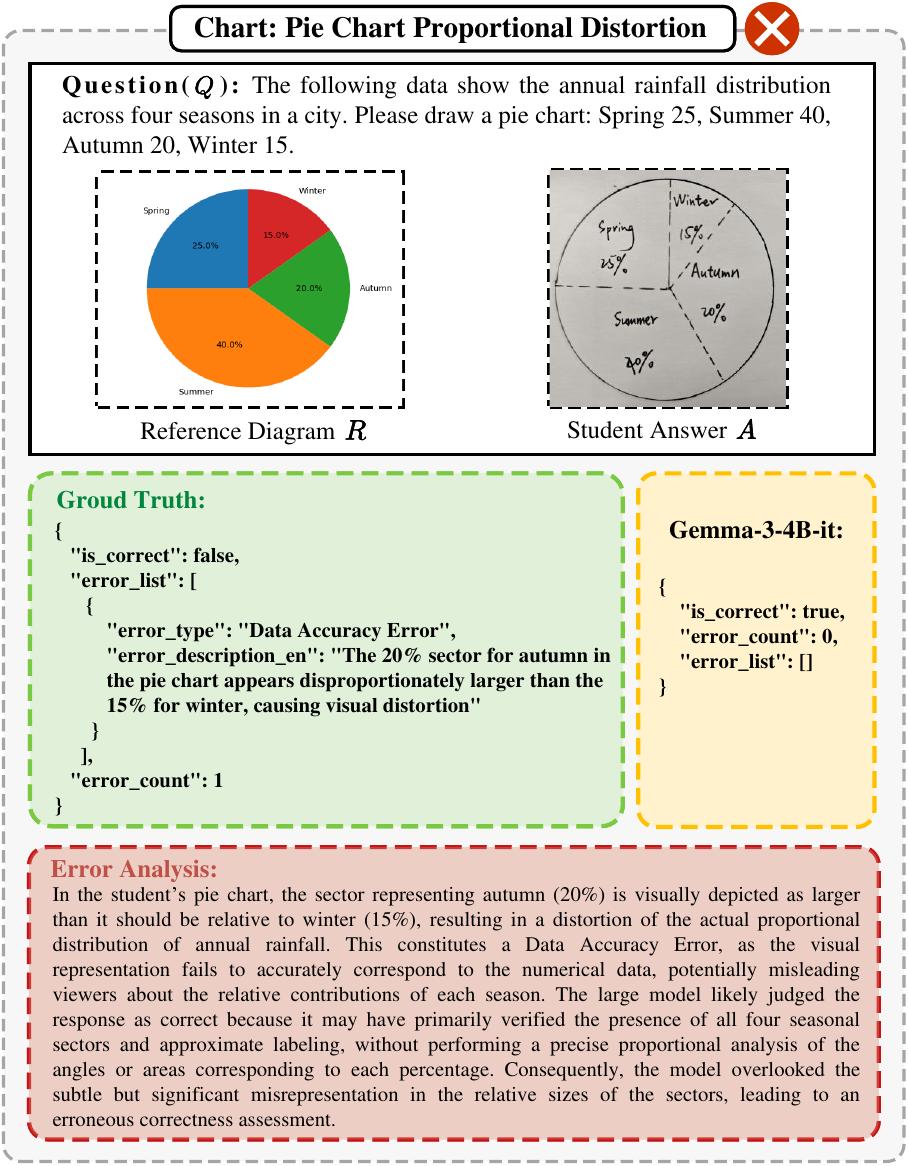}
    \caption{A sample error case in chart.}
\end{figure*}

\begin{figure*}[htbp]
    \centering
    \includegraphics[width=\linewidth]{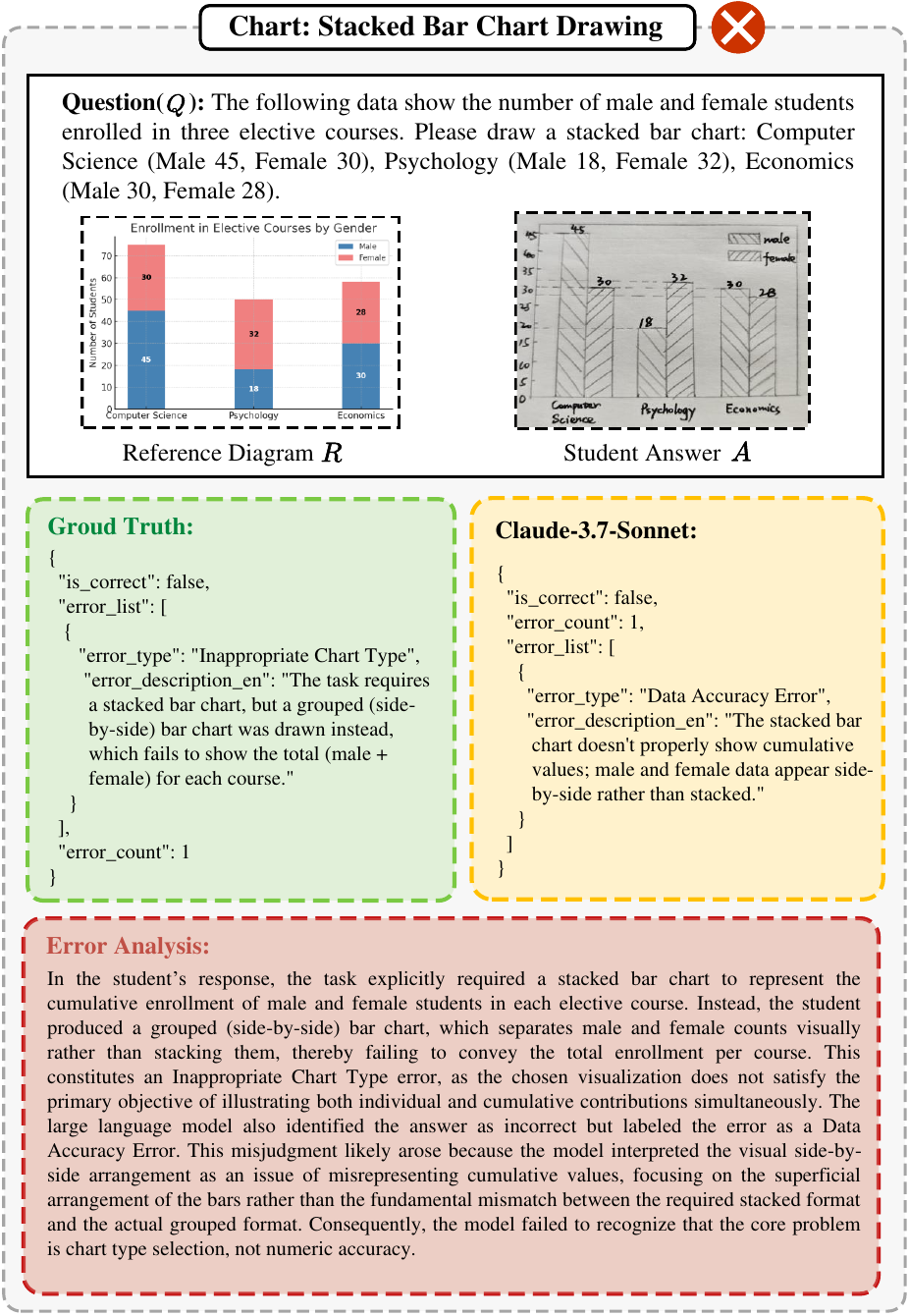}
    \caption{A sample error case in chart.}
\end{figure*}

\begin{figure*}[htbp]
    \centering
    \includegraphics[width=\linewidth]{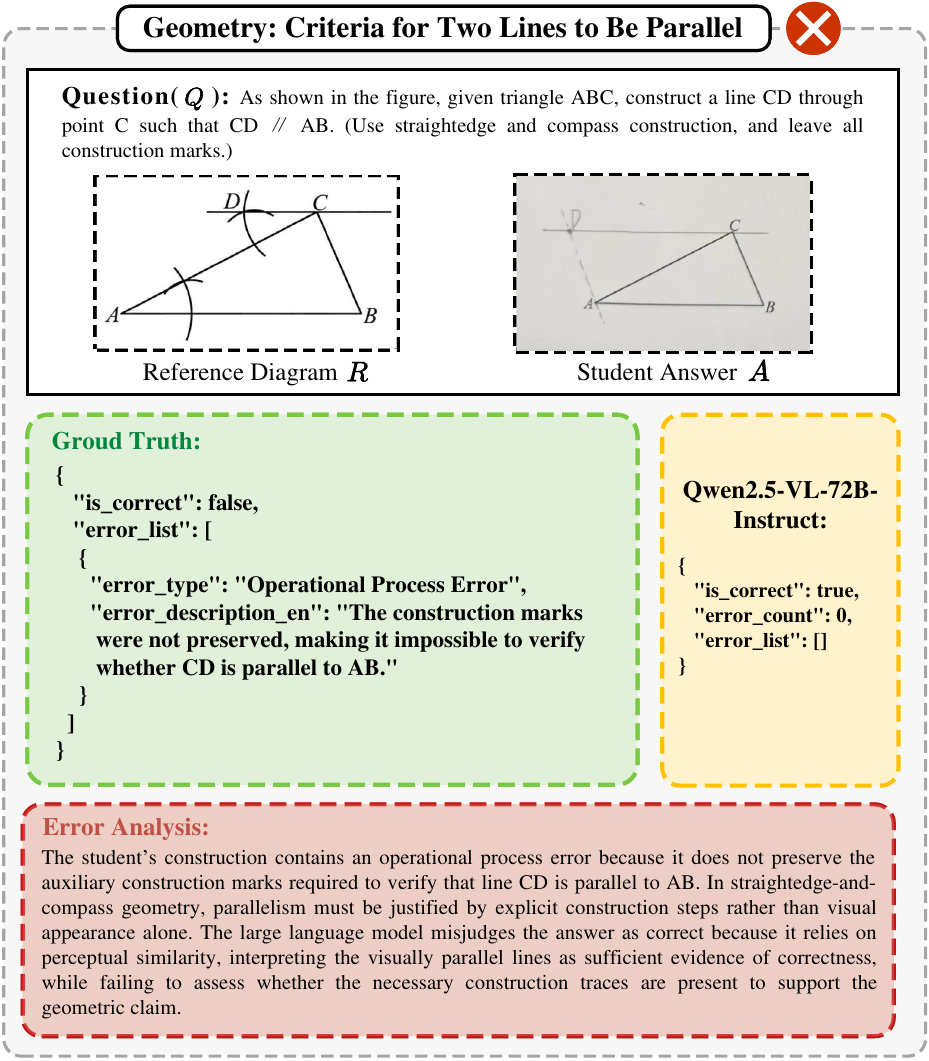}
    \caption{A sample error case in geometry.}
\end{figure*}

\begin{figure*}[htbp]
    \centering
    \includegraphics[width=\linewidth]{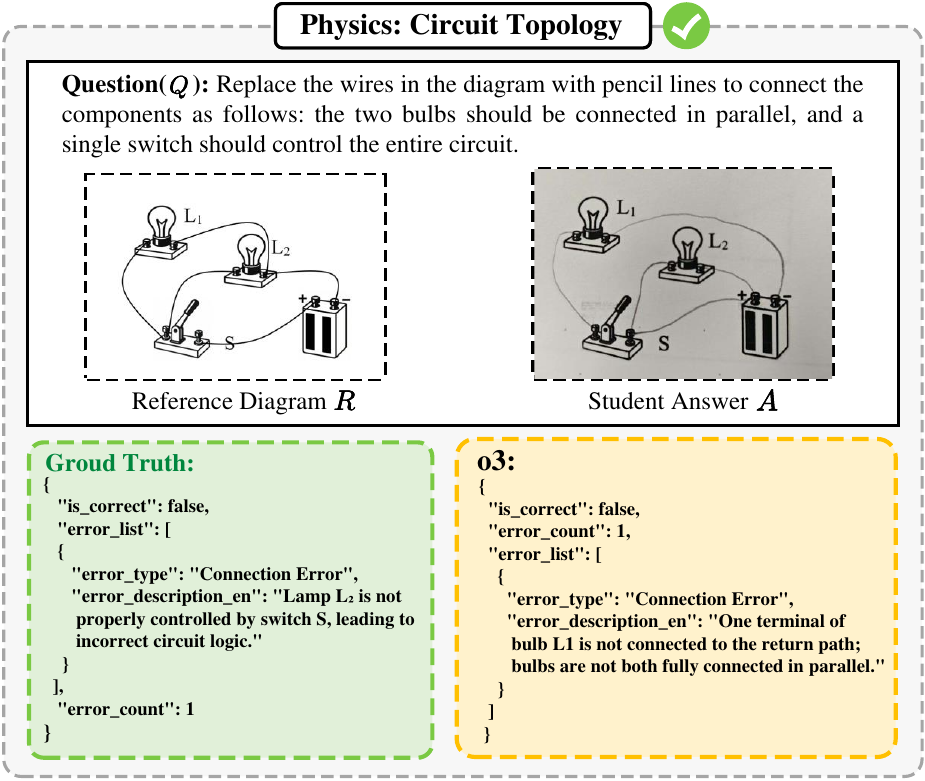}
    \caption{A sample correct case in physics.}
\end{figure*}

\end{document}